# "Explainable Machine Learning Analysis of Temperature Effects on Mechanical Strength of Fiber-Reinforced High-Performance Concrete"

Jagaran Chakma[a], Zhiguang Zhou[a*], Badhan Chakma[b], Priyanta Chakma[b]
[a] State Key Laboratory of Disaster Reduction in Civil Engineering, Tongji University, Shanghai 200092, China
[b] School of Civil Engineering, Chongqing Jiaotong University, Chongqing 400074, China

## Abstract

Predicting the compressive strength of fiber-reinforced high-performance concrete subjected to temperature exposure is challenging because to the intricate interdependencies of mixture composition, fiber properties, curing conditions, and temperature. A data-driven framework to predict compressive strength utilizing 1,204 experimental records containing a diverse array of material designs and exposure situations. This study evaluates eight machine-learning models and demonstrate that ensemble approaches routinely surpass individual learners. Both gradient boosting and extreme gradient boosting exhibit superior performance on unseen data, each attaining a R^2 of 0.903. Gradient boosting produces the minimal root mean square error (6.67 MPa), whereas extreme gradient boosting achieves the lowest mean absolute error (3.97 MPa) and mean absolute percentage error (10.60%). The models are further interpreted utilizing Shapley additive explanations and partial dependence analysis, which reveal temperature as the primary factor influencing compressive strength, succeeded by fine aggregate, superplasticizer, cement content, water-to-binder ratio, and water content. Strength diminishes significantly with rising temperature, albeit the extent of deterioration is contingent upon the specific combination design. The best performing models are subsequently implemented in a Python-based graphical user interface for expedited prediction. Our findings offer a precise and comprehensible methodology for evaluating temperature-dependent compressive strength and facilitate the practical, data-driven design of fiber-reinforced high-performance concrete.

∗Corresponding author: Zhiguang Zhou,
E-mail addresses:
chakmajagaran@tongji.edu.cn (J. Chakma),
zgzhou@tongji.edu.cn (Z. Zhou),
badhanchakma737@gmail.com (B. Chakma)
priyantachakma11@gmail.com (P. Chakma)

## 1. Introduction

High-performance concrete (HPC) is widely used in modern construction because of its superior mechanical strength, high durability, low permeability, and improved resistance to aggressive environmental conditions [1]. These advantages make HPC highly suitable for high-rise buildings, bridges, offshore structures, and other demanding engineering applications [2]. However, HPC also has some limitations, such as increased brittleness, lower ductility, higher material cost, and greater sensitivity to thermal exposure and microcracking under extreme conditions [3], [4]. Among its mechanical properties, compressive strength is one of the most important parameters used to evaluate its overall performance [5], since it directly affects structural safety, serviceability, and material selection [6], [7]. In fiber-reinforced high-performance concrete, compressive strength is influenced by many factors, including cement content, aggregate composition, water-to-binder ratio, superplasticizer dosage, fiber characteristics, curing age, and specimen dimensions [8], [9]. When HPC is exposed to different temperature conditions, the prediction of compressive strength becomes even more challenging [10] because temperature can significantly alter the microstructure, bonding condition, and overall mechanical response of the material [11], [12].

To address the limitations of conventional empirical and regression-based methods, machine learning (ML) techniques have been increasingly applied to predict the compressive strength of concrete materials [13], [14]. Models such as decision tree, support vector machine, artificial neural network, random forest, gradient boosting, LightGBM, and XGBoost have demonstrated strong capability in capturing complex nonlinear relationships among input variables [15-22]. These approaches significantly reduce resilience on extensive cost and time-consuming laboratory testing [23], [24]. Previous studies have reported that ML models can provide reliable and efficient prediction of compressive strength for different types of concrete materials [25], [26].

However, a thorough examination of the current literature reveals that numerous significant issues remain unaddressed. Initially, numerous prior investigations were conducted utilizing somewhat limited datasets. Despite frequent reports of excellent prediction accuracy, such outcomes do not invariably ensure robust generalization capability [27-29]. Secondly, certain research examined just a restricted set of input variables, despite the fact that the compressive strength of fiber-reinforced high-performance concrete (HPC) is influenced by multitudinous interrelated aspects pertaining to mixture design, fiber characteristics, curing conditions, and specimen shape [30], [31]. Third, in research concerning heat exposure, the predictive models were frequently developed for particular materials, restricted temperature ranges, or specific experimental settings, so diminishing their broader application [32-35]. Moreover, several earlier studies primarily emphasized predicted accuracy, whilst relatively less focus was directed towards model interpretability and practical implementation [36].

Therefore, a clear research gap exists in the current literature. There is still a need for a more comprehensive ML-based framework that uses a relatively large and diverse database, includes a broad range of temperature conditions, compares both single and ensemble learning models, and explains the prediction behavior of the best-performing models in an interpretable manner. Moreover, practical tools that allow engineers and researchers to directly use the developed models are still limited in existing research.

To address these limitations, the present study develops a machine learning framework for predicting the compressive strength of fiber-reinforced high-performance concrete under varying temperature conditions. A total of 1204 data samples with 17 input variables were collected from published experimental studies. These variables include mixture composition, fiber-related parameters, temperature, curing age, and specimen dimensions. In this work, eight ML models were evaluated, including four single models and four ensemble models, to identify the most suitable prediction approach.

The main contribution of this study is the development of accurate interpretable and practically deployable ML framework for temperature-dependent fiber-reinforced high-performance concrete prediction. The findings of this study are expected to provide useful support for material design, strength estimation, and engineering decision-making.

Table 1 Statistical summary of the inputs and Compressive strength.

| Response | Parameter | Count | Mean | Std. | Min | 25% | 50% | 75% | Max |
|---|---|---|---|---|---|---|---|---|---|
| Compressive Strength (MPa) | Cement (Kg/m$^3$) | 1204 | 430.53 | 93.41 | 250 | 398 | 403 | 465.5 | 1021 |
| | FAgg (Kg/m$^3$) | 1204 | 852.67 | 268.39 | 373 | 689 | 765 | 1055 | 1345 |
| | CSA (Kg/m$^3$) | 1204 | 917.10 | 220.75 | 115 | 765 | 974.25 | 1057.5 | 1681 |
| | Water (Kg/m$^3$) | 1204 | 191.38 | 49.86 | 123 | 161 | 185 | 197 | 672 |
| | River sand (Kg/m$^3$) | 1204 | 793.39 | 298.88 | 343 | 668 | 750 | 934.75 | 1460 |
| | Silica fume (Kg/m$^3$) | 1204 | 31.92 | 26.68 | 7.5 | 16.69 | 16.9 | 40 | 154 |
| | Superplasticizer (Kg/m$^3$) | 1204 | 42.41 | 177.96 | 0 | 4 | 4.7 | 6.6 | 1030 |
| | W/B (Kg/m$^3$) | 1204 | 0.44 | 0.14 | 0.18 | 0.37 | 0.4 | 0.5 | 0.88 |
| | Type of fibers | Basalt, Carbon, Glass, Steel, Polypropylene, Non-fiber | | | | | | | |
| | Fibers (%) | 1204 | 0.84 | 0.83 | 0 | 0.13 | 0.64 | 1 | 4 |
| | Aspect ratio (%) | 1204 | 104.24 | 163.71 | 0 | 52 | 58.33 | 62.5 | 400 |
| | Temperature ℃ | 1204 | 297.28 | 291.85 | -60 | 20 | 200 | 600 | 1200 |
| | Curing age (Day) | 1204 | 26.57 | 11.23 | 3 | 28 | 28 | 28 | 90 |
| | Specimen width (cm$^3$) | 1204 | 10.65 | 1.96 | 0 | 10 | 10 | 10 | 15 |
| | Specimen length (cm$^3$) | 1204 | 10.65 | 1.96 | 0 | 10 | 10 | 10 | 15 |
| | Specimen height (cm$^3$) | 1204 | 11.54 | 3.31 | 0 | 10 | 10 | 10 | 20 |
| | Compressive strength (MPa) | 1204 | 40.17 | 22.50 | 20.2 | 23.4 | 32 | 50.9 | 133.6 |

## 2. Methodology

Figure 1 depicts the detailed methodological framework employed in this study, including dataset preparation, model creation, validation, interpretation, and application. It also encapsulates the amalgamation of ML modelling, SHAP/PDP analysis, comparison with prior studies, and web-based deployment.

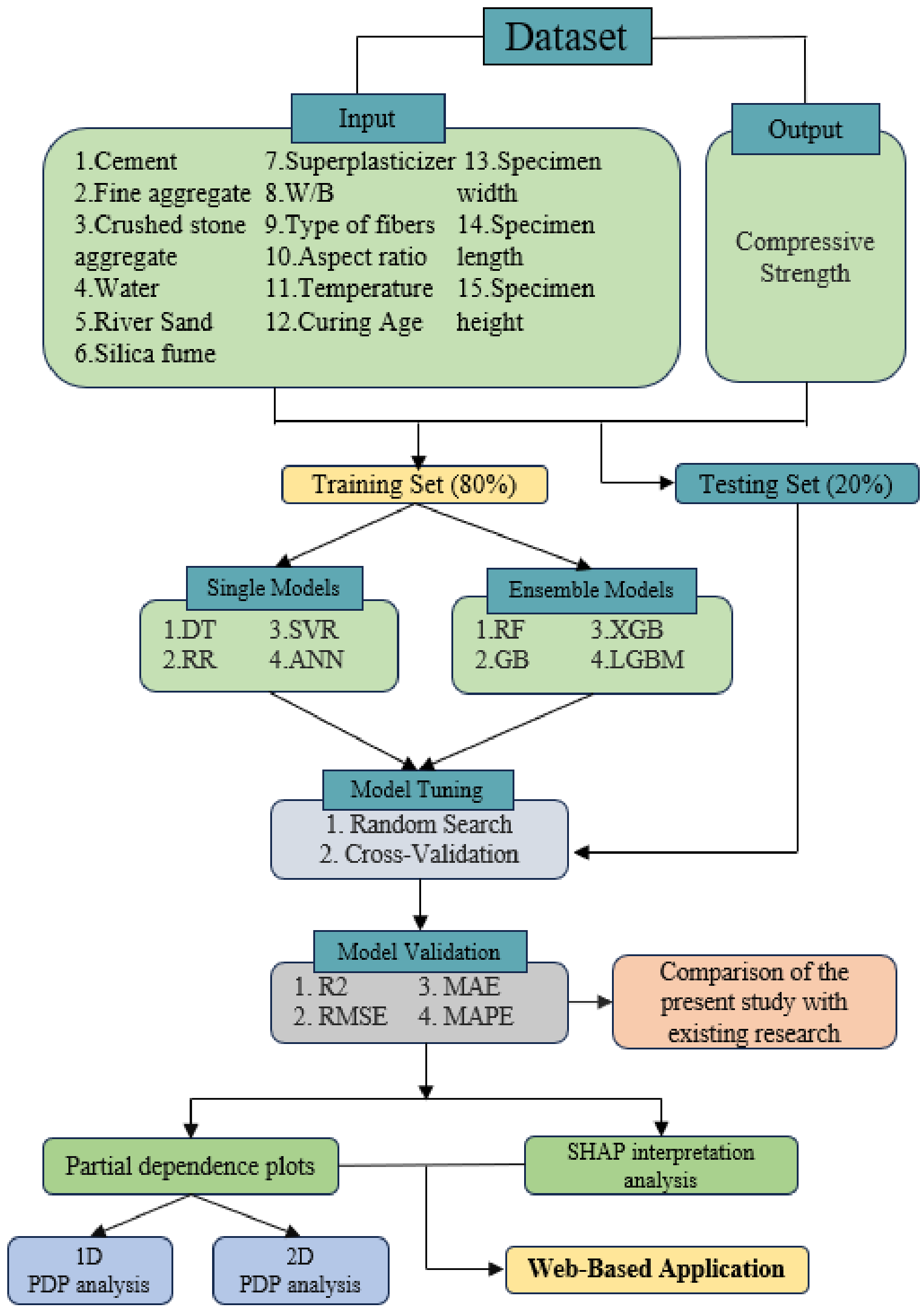

Fig. 1 Flowchart of the procedure in this study.

## 2.1 Data preprocessing

Data preprocessing was performed to rectify missing or zero values in the dataset before model building. In the compiled dataset, certain input variables included zero values indicative of missing or unavailable data rather than genuine measurements. The mean imputation method was employed to address this issue. Specifically, zero values in each feature were substituted with the mean value of the relevant variable derived from the available non-zero observations. This method maintains the statistical properties of the dataset while averting potential information loss that may arise from the removal of samples. Mean imputation preserves data continuity and mitigates any bias throughout the model training phase. Implementing this preprocessing step enhances the dataset's completeness and suitability for constructing the machine learning model aimed at predicting the compressive strength of fiber-reinforced high-performance concrete.

## 2.2 Statistical Analysis

Statistical analysis was used to characterize the dataset and examine linear connections between input variables and compressive strength before model creation. The dataset contained mixture proportions, fiber qualities, curing procedures, specimen size, and temperature exposure levels from published experimental research. Mixture composition, fiber-related factors, temperature, curing age, specimen size, and compressive strength were observed in 1204 [31], [44], [37-53] observations. Table 1 shows numerical variable descriptive statistics, whereas Figs. 2 and 3 show the Pearson correlation matrix and variable-compressive strength interactions. The dataset has wide variability, suggesting model development diversity. The compressive strength averaged 40.17 MPa and ranged from 20.2 to 133.6 MPa. The input variables varied greatly, including W/B ratio ranged from 0.18 to 0.88, temperature from -60 to 1200 °C, and curing time from 3 to 90 days. The dataset included basalt, carbon, glass, steel, polypropylene, and non-fiber specimens with various fiber content, aspect ratios, and specimen dimensions. The dataset is diverse and representative enough for robust compressive strength machine learning models, according to statistical analysis.

The Pearson correlation matrix indicates multiple significant linear relationships among the predictors. A perfect positive correlation ($r = 1.00$) was noted between specimen width and specimen length, suggesting that these two variables were equivalent for several samples and may yield redundant geometric information. Cement exhibited a significant negative correlation with W/B ($r = -0.60$), whereas water demonstrated a moderate positive correlation with W/B ($r = 0.55$). This is physically acceptable, as the water-to-binder ratio escalates with increased water content and diminishes with higher binder quantity. River sand demonstrated a moderate positive association with fine aggregate ($r = 0.59$), but fine aggregate exhibited moderate negative correlations with CSA ($r = -0.38$), silica fume ($r = -0.43$), and curing age ($r = -0.45$). The connections demonstrate a certain level of interaction among the mixture design variables, while the matrix does not imply significant multicollinearity among the majority of the predictors. The correlation study indicates that the target variable, compressive strength, exhibits the strongest positive connection with cement content ($r = 0.459$), followed by curing age ($r = 0.260$) and silica fume content ($r = 0.254$). These findings align with the function of cementitious materials and curing maturity in improving hydration and compacting the microstructure. The most pronounced negative connection with compressive strength was identified for the water-to-binder ratio (r = −0.404), indicating that an increased water content in respect to binder typically diminishes matrix density and strength. Additional variables exhibiting weak to moderate negative associations with compressive strength comprise fine aggregate (r = −0.233), river sand (r = −0.204), temperature (r = −0.203), specimen width (r = −0.186), specimen length (r = −0.186), and CSA (r = −0.153). The relationship between compressive strength and fiber content (r = −0.095), water (r = −0.058), superplasticizer (r = 0.046), and aspect ratio (r = 0.030) was notably weak, indicating that their influences may be nonlinear or contingent upon interactions with other variables rather than direct linear correlations.

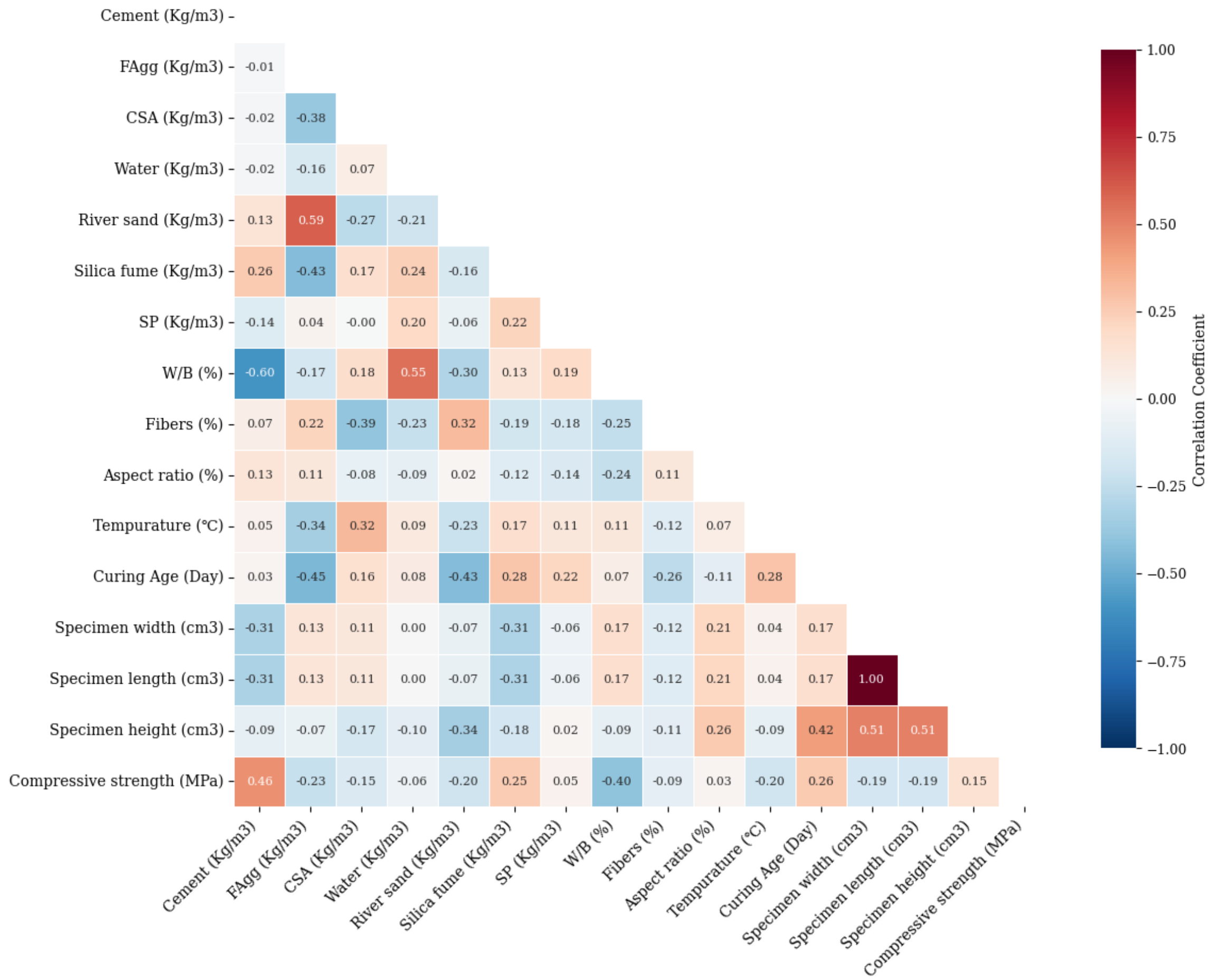

Fig 2. Pearson correlation plot for the compressive strength.

The statistical analysis demonstrates that the literature-based dataset is adequately diverse and exhibits significant variation in mixture proportions, specimen geometry, curing conditions, fiber properties, and temperature exposure. The Pearson correlation results indicate that compressive strength is predominantly affected by binder-related parameters, curing age, temperature, and water-to-binder ratio, whereas others variables exert relatively minimal linear impacts. The

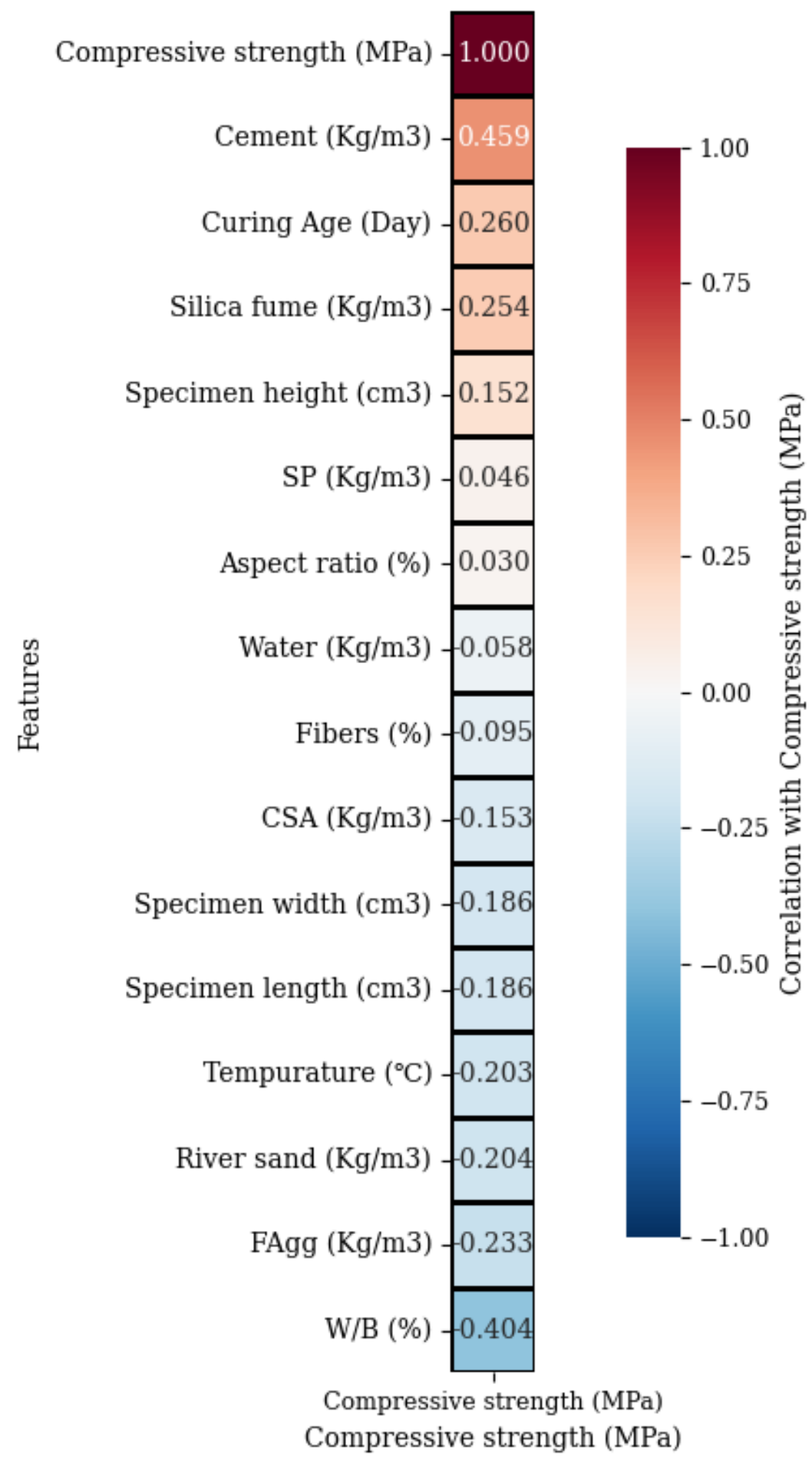

Fig 3. Correlation of Mix Parameters with Compressive Strength.

Pearson correlation, which only measures linear dependence, indicates that moderate correlations do not inherently suggest low predictive significance in machine learning models.

The dataset for predictive modelling was partitioned into training and testing subsets utilizing an 80:20 ratio across all models. Consequently, 80% of the data was allocated for model training, and the remaining 20% was designated for independent testing and performance assessment. This cohesive data-splitting approach enabled a uniform and equitable comparison across all constructed models.

### 2.3 Machine learning models and evaluation metrics

According to the statistical analysis, the dataset was configured for machine learning predictions regarding the compressive strength of fiber-reinforced HPC under different temperature contexts. The input variables included factors related to mixture composition, such as cement, fine aggregate, CSA, water, river sand, silica fume, superplasticizer, and water-to-binder ratio, as well as fiber-related parameters including fiber content, fiber type, and aspect ratio. Furthermore, curing age, temperature, and specimen dimensions were incorporated as explanatory variables, while compressive strength was designated as the goal output. The descriptive statistics and correlation analysis indicated significant variability and meaningful relationships among the variables; therefore, both single learner models and ensemble learner models were utilized to elucidate the underlying mapping between the input features and compressive strength. This study implemented eight machine learning models. The individual models comprised Decision Tree (DT), Ridge Regression (RR), Support Vector Regression (SVR), and Artificial Neural Network (ANN). The ensemble models comprised Random Forest (RF), Gradient Boosting (GB), Light Gradient Boosting Machine (LightGBM), and Extreme Gradient Boosting (XGBoost). These models were chosen as they exemplify several learning philosophies, including linear regularized learning, nonlinear kernel learning, tree-based partitioning, neural computation, and ensemble boosting/bagging techniques. Their joint application facilitates a thorough comparison between simple predictors and more sophisticated nonlinear models.

### 2.3.1 Single machine learning models

(1) Decision tree

A Decision Tree is a nonparametric supervised learning method that predicts outcomes by recursively dividing the feature space into smaller segments. At each node, the dataset is partitioned based on the feature and threshold that reduce the prediction error to the greatest extent. In regression, the predicted compressive strength at a terminal leaf is the mean of the training responses within that area. The forecast can be articulated as

$$\hat{f}_c(z) = \sum_{m=1}^{M} \omega_m \, \mathbb{I}\,(z \in \delta_m) \tag{1}$$

$\hat{f}_c(z)$ represents the predicted compressive strength, z signifies the input feature vector, $\delta_m$ denotes the *m*-th partitioned region, $\omega_m$ indicates the mean compressive strength of the training samples within that region, and $\mathbb{I}\,(\cdot)$ is an indicator function that equals 1 when the condition is met and 0 otherwise. Decision trees are beneficial as they can contain nonlinear interactions and are easily interpretable; yet, they may be prone to overfitting when utilized as independent learners.

(2) Ridge regression

Ridge Regression is a regularized linear model that quantifies the link between predictors and compressive strength, imposing a penalty on the magnitude of the regression coefficients. This regularization diminishes model variance and enhances generalization when the predictors exhibit correlation. The model is expressed as

$$\hat{f}_c = \alpha_0 + \sum_{j=1}^{p} \alpha_j q_j \tag{2}$$

When $\alpha_0$ represents the intercept, $\alpha_j$ denotes the regression coefficients, $q_j$ signifies the input variables, and $p$ indicates the number of predictors. The coefficients are derived by minimising the subsequent penalized loss function:

$$\tau(\alpha) = \sum_{i=1}^{N}\left(y_i - \alpha_0 - \sum_{j=1}^{p} \alpha_j q_{ij}\right)^2 + \lambda \sum_{j=1}^{p} \alpha_j{}^2 \quad (3)$$

where $y_i$ represents the observed compressive strength for the $i$-th sample, $N$ denotes the number of observations, and $\lambda$ signifies the regularization parameter. Ridge Regression serves as an effective baseline model due to its robustness and computing efficiency; yet, its predictive power is constrained in scenarios where variable relationships exhibit substantial nonlinearity.

(3) Support vector regression

Support Vector Regression is a kernel-based learning approach that aims to identify a function exhibiting greatest flatness while maintaining prediction errors within a specified tolerance. In contrast to conventional linear regression, Support Vector Regression (SVR) may represent nonlinear relationships via kernel transformation. The regression function can be articulated as

$$\hat{f}_c(u) = \sum_{i=1}^{N_s}(\beta_i - \beta_i^*)K(u_{i,} u) + \delta \quad (4)$$

Here, $N_s$ denotes the quantity of support vectors, $\beta_i$ and $\beta_i^*$ represent the Lagrange multipliers, $K\left(u_{i,} u\right)$ signifies the kernel function, $u_i$ indicates the support vector, $u$ refers to the input vector, and $\delta$ is the bias term. The radial basis function kernel is typically denoted as

$$K\left(u_{i,} u\right) = exp(-\eta||u_i - u)^2 \quad (5)$$

where $\eta$ regulates the kernel width. SVR is very adept at handling complicated datasets with nonlinear relationships and can yield precise predictions when its hyperparameters are appropriately optimized.

(4) Artificial neural network

An Artificial Neural Network is a computational model derived from biological neurons, proficient at learning complex nonlinear input-output interactions. In a feedforward artificial neural network, each neuron calculates a weighted sum of the incoming signals and subsequently applies an activation function. The anticipated compressive strength for a network with a single hidden layer may be articulated as

$$\hat{f}_c(x) = \psi_0\left(\sum_{r=1}^{H} v_r \psi_h\left(\sum_{j=1}^{p} \omega_{rj} x_j + b_r\right) + b_0\right) \quad (6)$$

In this context, $x_j$ represents the $j$-th input variable, $\omega_{rj}$ signifies the weight linking the $j$-th input to the $r$-th hidden neurone, $v_r$ denotes the weight connecting the hidden neurone to the output layer, while $b_r$ and $b_0$ are bias terms. $H$ indicates the total number of hidden neurons, $\psi_h(\cdot)$ refers to the activation function of the hidden layer, and $\psi_0(\cdot)$ pertains to the output activation function. Artificial Neural Networks (ANN) can approximate intricate nonlinear functions, making them highly effective for forecasting compressive strength based on multiple material and thermal variables.

### 2.3.2 Ensemble machine learning models

(1) Random forest

Random Forest is an ensemble variation of the decision tree model. It generates several regression trees utilizing bootstrap samples from the training dataset and subsequently consolidates their predictions. The ultimate prediction is derived by averaging the outputs of all trees.

$$\hat{f}_c^{RF}(s) = \frac{1}{B}\sum_{b=1}^{B} T_b(s) \qquad (7)$$

$B$ represents the quantity of trees in the forest, $T_b(s)$ is the prediction from the $b$-th tree, and s signifies the input vector. Random Forest enhances prediction stability and mitigates overfitting relative to an individual decision tree, as each tree is trained on a distinct bootstrap sample and a random selection of attributes.

(2) Gradient boosting

Gradient Boosting is a sequential ensemble method wherein weak learners, typically shallow decision trees, are incrementally incorporated to diminish the residual errors of prior learners. The model can be articulated as

$$\hat{f}_c^{GB}(r) = \sum_{m=0}^{M} v h_m(r) \qquad (8)$$

where $h_m(r)$ denotes the $m$-th weak learner, $M$ represents the number of boosting iterations, and $v$ signifies the learning rate that regulates the contribution of each learner. At each level, the new learner is aligned with the negative gradient of the loss function, enabling the ensemble to systematically reduce prediction error. Gradient Boosting is proficient for intricate nonlinear regression challenges and typically attains robust predictive efficacy.

(3) LightGBM

Light Gradient Boosting Machine is an efficient gradient boosting system optimized for rapid training and reduced memory consumption. It employs a leaf-wise tree development technique that enhances predicted accuracy by concentrating on the leaf that yields the most significant reduction in loss. The prediction can be expressed generally as

$$\hat{f}_c^{LGB}(m) = \sum_{\mathrm{k}=1}^{K} g_{\mathrm{k}}(m) \quad (9)$$

where $g_{\mathrm{k}}(m)$ represents the output of the $k$-th boosted tree and $K$ signifies the total number of trees. LightGBM distinguishes itself from traditional boosting through its histogram-based decision process and enhanced split optimisation, rendering it especially appealing for extensive and diverse datasets. The current dataset comprises a combination of material, geometric, and thermal factors, suggesting that LightGBM will proficiently identify intricate nonlinear relationships.

(4) XGBoost

XGBoost is a sophisticated gradient boosting technique that integrates regularization, second-order optimization, and effective management of sparse and structured data. The model's prediction is denoted as

$$\hat{f}_c^{XGB}(a) = \sum_{\mathrm{t}=1}^{T} \varphi_{\mathrm{t}}(a), \qquad \varphi_t \in \mathcal{F} \quad (10)$$

where $\varphi_{\mathrm{t}}(a)$ represents the output of the $t$-th regression tree, $T$ signifies the total number of trees, and $\mathcal{F}$ symbolises the functional space including all potential regression trees. The objective function of XGBoost is typically articulated as

$$\mathcal{L} = \sum_{i=1}^{N} \ell(y_{i,}\hat{y}_i) + \sum_{t=1}^{T} \Omega(\phi_t) \quad (11)$$

where $\ell(y_{i,}\hat{y}_i)$ represents the training loss and $\Omega(\phi_t)$ is the regularisation term that governs tree complexity. Due to its regularisation process and effective boosting framework, XGBoost has emerged as one of the most dependable algorithms for tabular engineering datasets.

### 2.3.3 Model evaluation metrics

Various statistical assessment measures were utilized to evaluate the predictive performance of the created models. These measures measure the discrepancy between anticipated and experimental compressive strengths and offer additional insights into model accuracy, error magnitude, and goodness of fit.

The coefficient of determination ($R^2$) was employed to assess the proportion of variance in compressive strength elucidated by the model:

$$R^2 = 1 - \frac{\sum_{i=1}^{N}(y_i - \hat{y}_i)^2}{\sum_{i=1}^{N}(y_i - \bar{y})^2} \quad (12)$$

where $y_i$ represents the measured compressive strength, $\hat{y}_i$ denotes the projected compressive strength, and $\bar{y}$ signifies the mean of the recorded values. An elevated $R^2$ signifies superior explanatory efficacy.

The mean absolute error (MAE) was employed to quantify the average absolute variation between projected and actual values:

$$MAE = \frac{1}{N}\sum_{i=1}^{N}|y_i - \hat{y}_i| \tag{13}$$

The root mean square error (RMSE) was computed to highlight significant prediction errors:

$$RMSE = \sqrt{\frac{1}{N}\sum_{i=1}^{N}(y_i - \hat{y}_i)^2} \tag{14}$$

Furthermore, the mean absolute percentage error (MAPE) was employed to assess relative prediction accuracy in percentage terms:

$$MAPE = \frac{100}{N}\sum_{i=1}^{N}\left|\frac{y_i - \hat{y}_i}{y_i}\right| \tag{15}$$

A diminished value of MAE, RMSE, and MAPE signifies enhanced prediction accuracy, while an elevated $R^2$ denotes a more robust correlation between measured and anticipated compressive strengths.

As outlined in Section 2.2, the dataset was partitioned into 80% training data and 20% testing data, with this division applied uniformly across all individual and ensemble models to provide a fair and consistent comparison. The training subset was utilized to determine the predictive correlations between the input parameters and compressive strength, whereas the testing subset was employed for independent validation of the model's generalization capacity.

## 3. Results and discussion

### 3.1 Hyperparameter optimization

The hyperparameters of all generated machine learning models were optimized via the random search optimization method, which is computationally efficient and appropriate for exploring an extensive search space. The ideal hyperparameter values for the compressive strength response are displayed in Table 2. The findings indicate that each model necessitated a distinct parameter configuration to get optimal performance. The decision tree exhibited optimal performance with modest depth and regulated split parameters, signifying the necessity to curtail excessive model complexity. Ridge regression required a comparatively elevated regularization coefficient, indicating that regularity was essential for stabilizing the linear model. The chosen parameters of C, gamma, and epsilon for SVR demonstrate that the model used a flexible nonlinear fitting capability with regulated error tolerance.

Table 2 Optimized values of the hyperparameters for the compressive strength response.

| Response | Model | Hyperparameter optimization | Optimal value |
|---|---|---|---|
| Compressive strength | Decision tree | Max depth | 10 |
| | | Max. of features | sqrt |
| | | Min. of sample split | 5 |
| | | Min. sample leaf | 2 |
| | Ridge regression | Best alpha regularization | 19.31 |
| | Support vector regression | Gamma | auto |
| | | Epsilon | 0.119 |
| | | C | 88.59 |
| | Artificial neural network | Solver | adam |
| | | Max. of iteration | 1500 |
| | | Learning rate_init | 0.001 |
| | | Hidden layer sizes | (100,) |
| | | Batch size | 64 |
| | | Alpha | 0.0001 |
| | | Activation | relu |
| | Random forest | No. of estimators | 500 |
| | | Min. samples split | 2 |
| | | Min. samples leaf | 1 |
| | | Max. features | Log2 |
| | | Max. depth | None |
| | | Bootstrap | True |
| | Gradient boosting | Subsample | 1.0 |
| | | No. of estimators | 200 |
| | | Min. samples split | 2 |
| | | Min. samples leaf | 4 |
| | | Max. of features | sqrt |
| | | Max. depth | 4 |
| | | Learning rate | 0.2 |
| | LightGBM | Subsample | 0.7 |
| | | Reg_lambda | 0.1 |
| | | Reg alpha | 0.01 |
| | | No. of leaves | 31 |
| | | No. of estimators | 300 |
| | | Max. depth | 20 |
| | | Learning rate | 0.1 |
| | | Col sample by tree | 0.6 |
| | XGBoost | Subsample | 0.1 |
| | | Reg lambda | 1 |
| | | Reg alpha | 0.1 |
| | | No. of estimators | 500 |
| | | Min. child weight | 7 |
| | | Max. depth | 50 |
| | | Learning rate | 0.05 |
| | | Gamma | 0.5 |
| | | Col sample by tree | 0.7 |

The artificial neural network attained optimal results utilizing a single hidden layer including 100 neurons, employing ReLU activation, the Adam optimizer, and a minimal learning rate, indicating that a straightforward neural network architecture was adequate for this issue. Among ensemble approaches, random forest, gradient boosting, LightGBM, and XGBoost necessitate numerous estimators and supplementary regularity or sampling controls, illustrating the intricate nonlinear relationship between input variables and compressive strength. The optimized hyperparameters indicate that nonlinear and ensemble-based models are more appropriate than linear models for accurately representing the dataset's behavior.

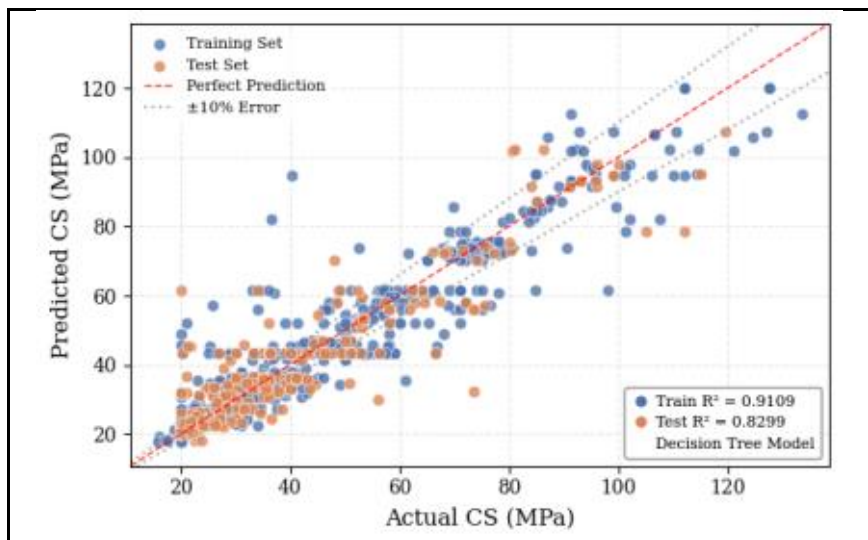

(a) Decision tree

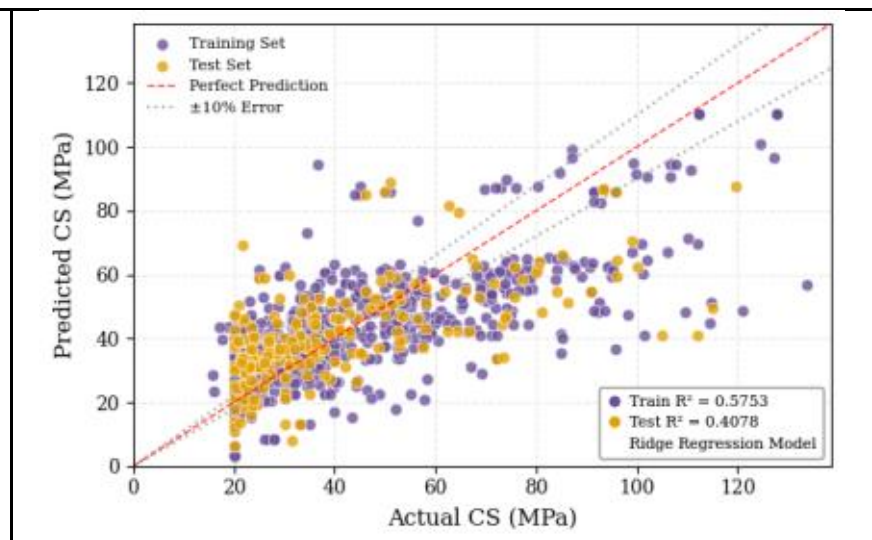

(b) Ridge regression

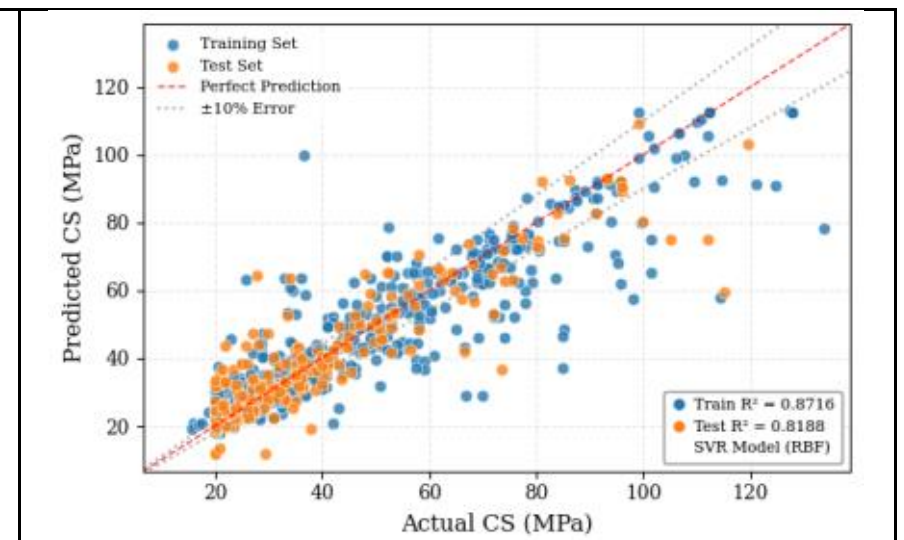

(c) Support Vector Regression

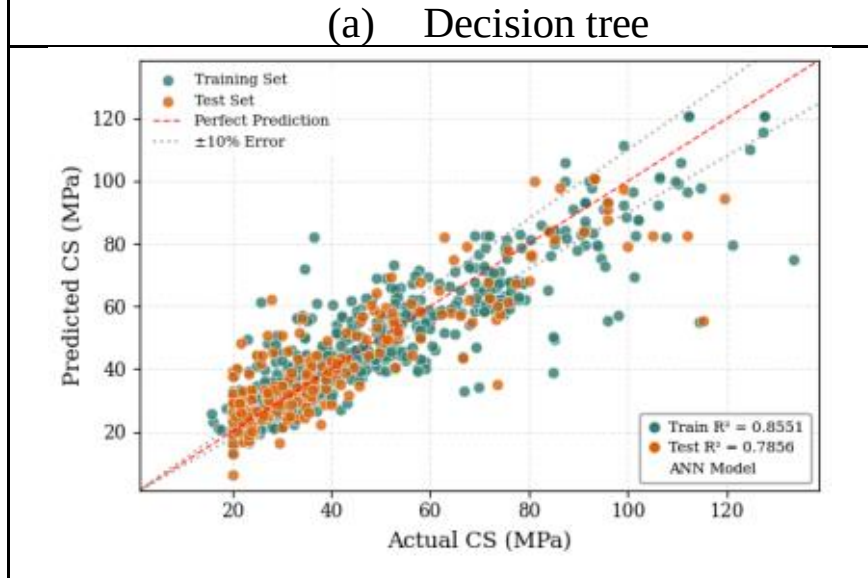

(d) Artificial neural network

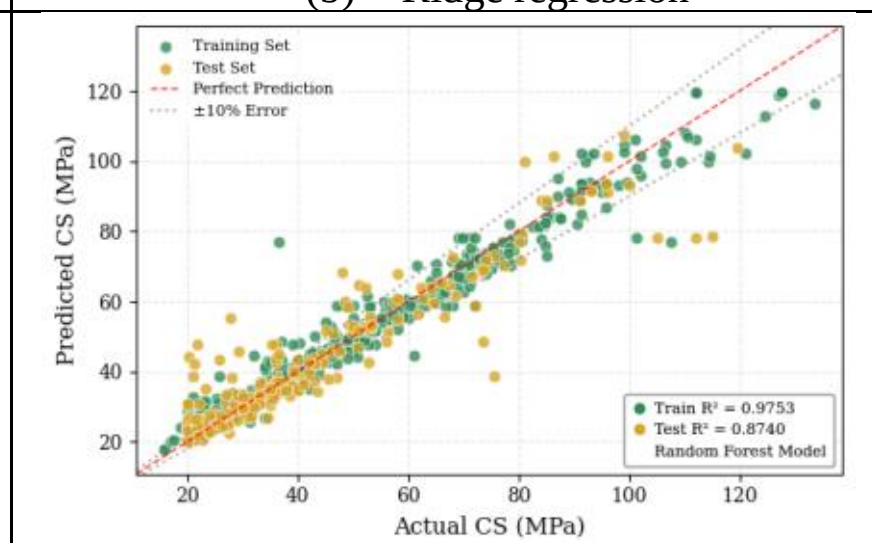

(e) Random forest

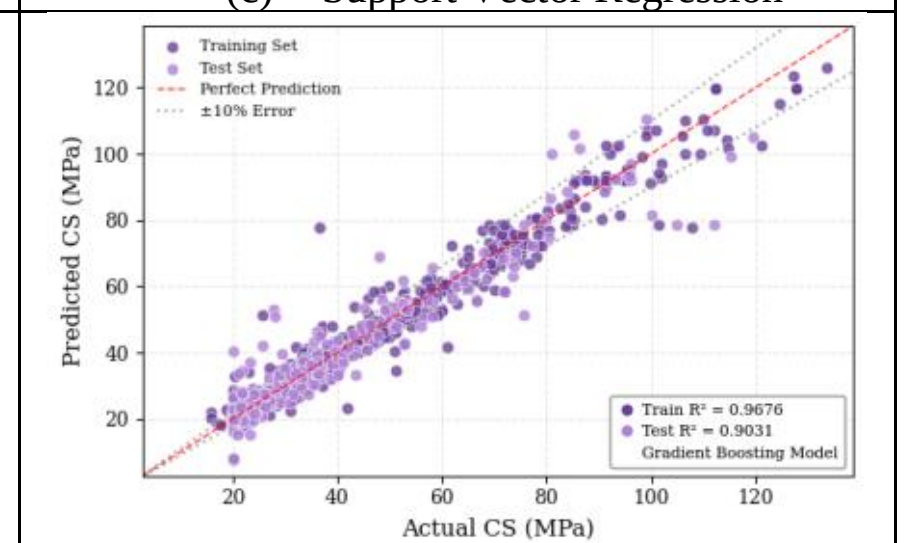

(f) Gradient boosting

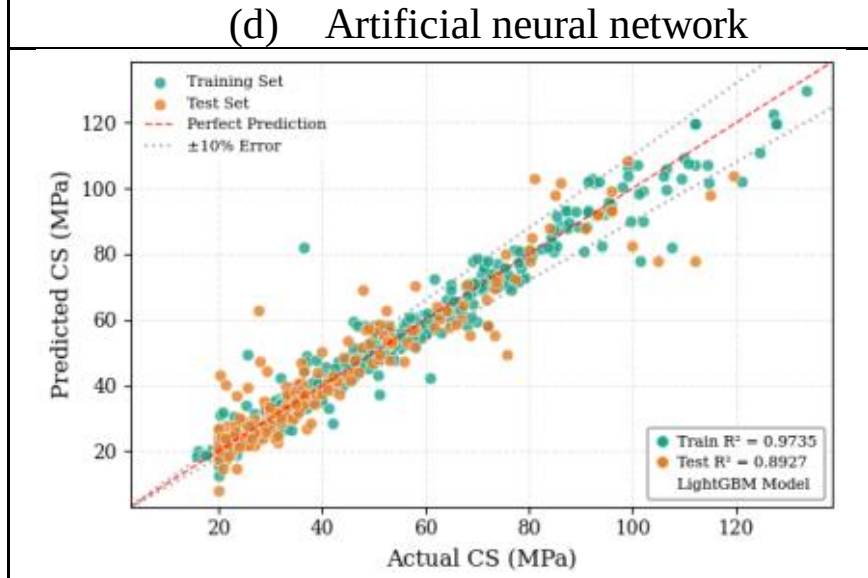

(g) Light GBM

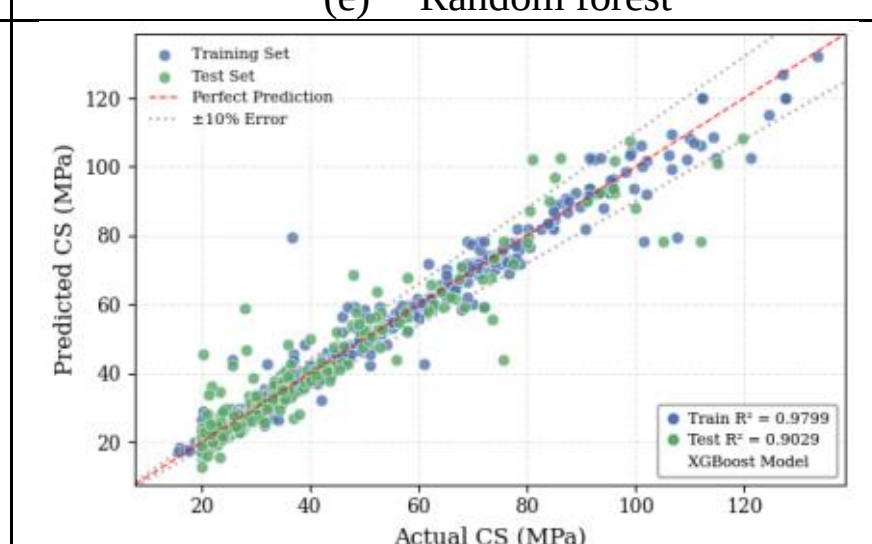

(h) XGBoost

Fig. 4 Predictive performance of ML models for compressive strength.

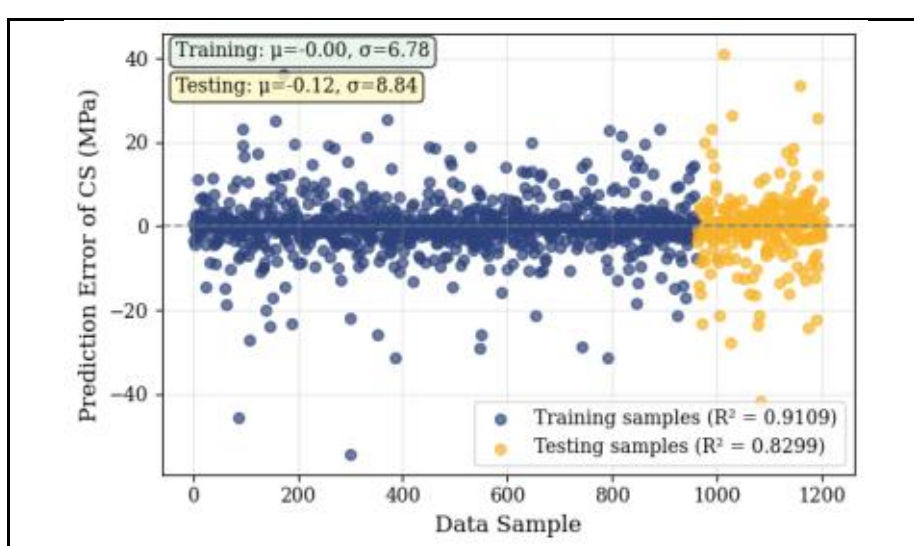

(a) Decision tree

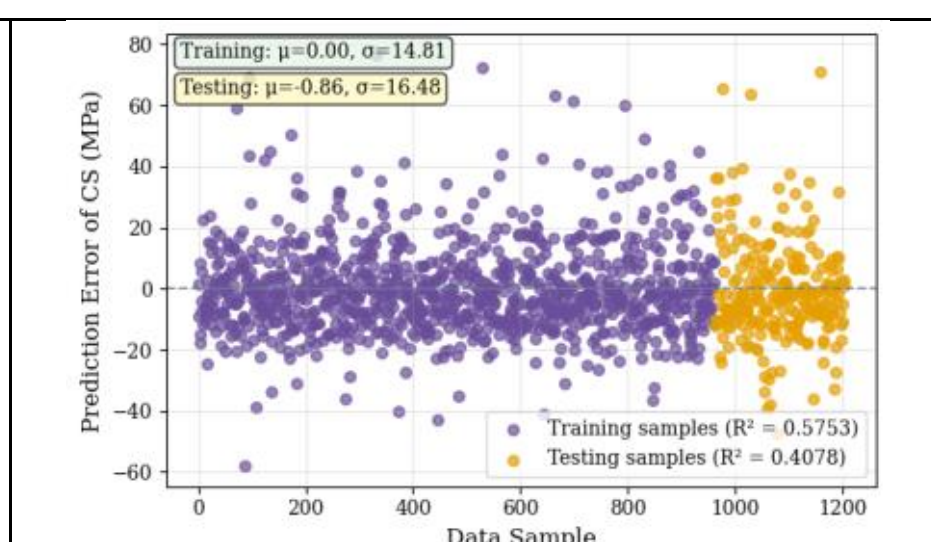

(b) Ridge regression

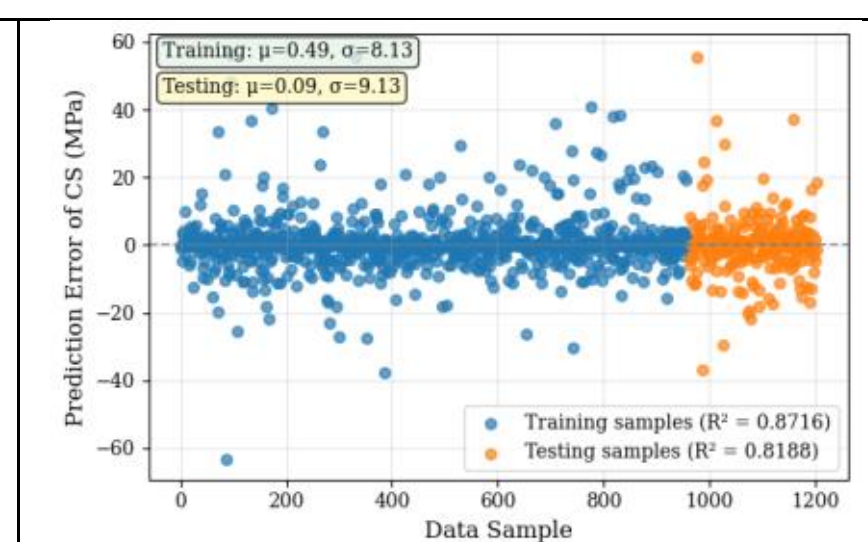

(c) Support Vector Regression

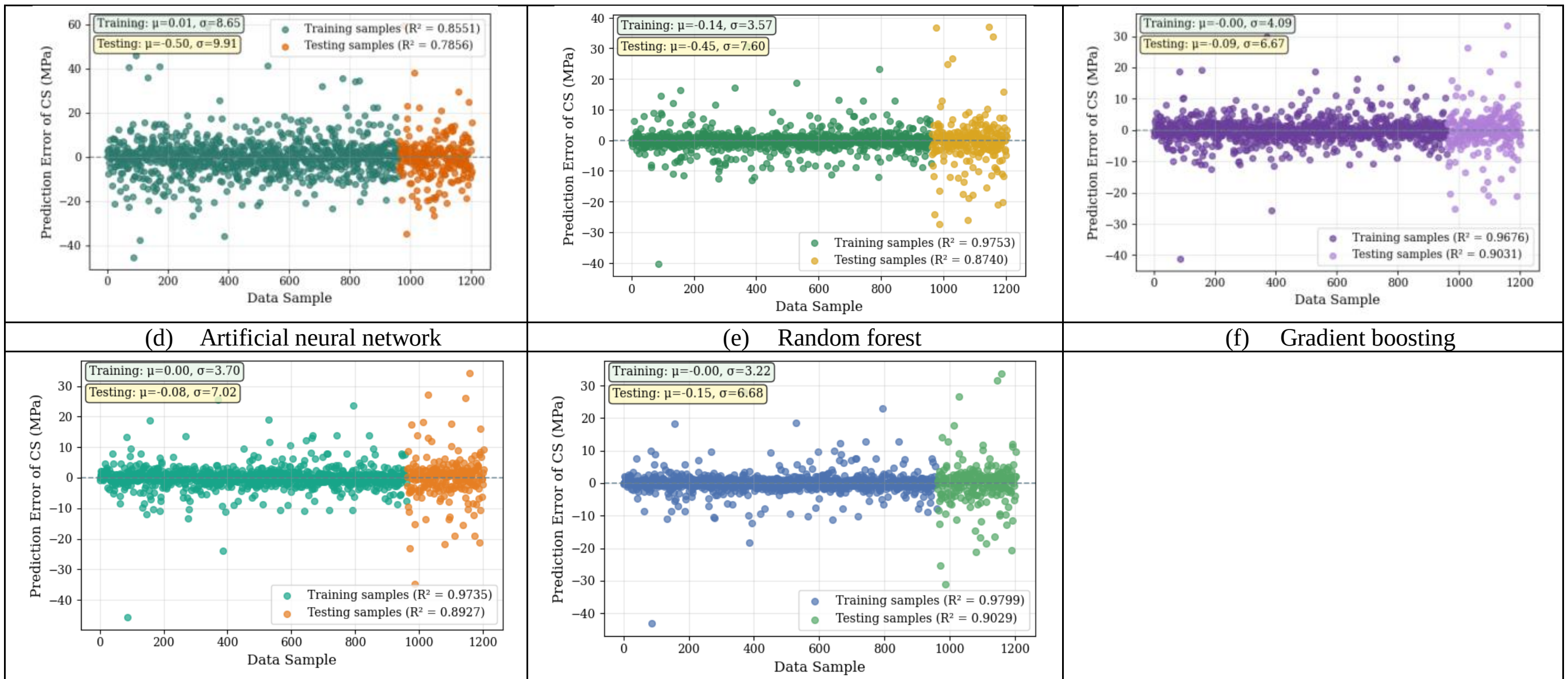

(d) Artificial neural network (e) Random forest (f) Gradient boosting

(g) Light GBM (h) XGBoost

Fig. 5 Residual plots of ML models for compressive strength.

## 3.2 Model performance evaluation

The predictive efficacy of the constructed models for compressive strength was evaluated utilizing the statistical metrics presented in Tables 3 and 4, along with the scatter and residual plots illustrated in Figs. 4 and 5 and the comparative metric plots in Fig.6. The ensemble-based models outperformed the individual and linear models. This is demonstrated by their reduced RMSE, MAE, and MAPE values, elevated R^2, tighter clustering around the ideal prediction line in the scatter plots, and more confined residual distributions around zero.

In the training dataset, random forest exhibited the greatest R^2 value of 0.98, whilst XGBoost recorded the lowest MAE of 1.61 and MAPE of 4.01. LightGBM with gradient boosting demonstrated robust predictive capability, with data points strongly aligned along the diagonal and residuals clustered around zero. Conversely, ridge regression exhibited the least favorable performance (RMSE = 14.81, R^2 = 0.58), characterized by significant scatter and extensive residual dispersion, suggesting that the linear model failed to sufficiently represent the nonlinear behavior of compressive strength.

Table 3 Statistical measures of the training datasets.

| Training set | | RMSE | MAE | MAPE | $R^2$ |
|---|---|---|---|---|---|
| New developed models | Decision tree | 6.78 | 4.02 | 10.46 | 0.91 |
| | Ridge regression | 14.81 | 10.88 | 29.81 | 0.58 |
| | Support vector regressor | 8.14 | 4.32 | 10.84 | 0.87 |
| | Artificial neural network | 8.65 | 5.68 | 15.13 | 0.79 |
| | Random forest | 3.57 | 4.49 | 4.78 | 0.98 |
| | Gradient bossting | 4.09 | 2.54 | 6.85 | 0.97 |
| | LightGBM | 3.71 | 2.05 | 5.33 | 0.97 |
| | XGBoost | 3.22 | 1.61 | 4.01 | 0.90 |

Table 4 Statistical measures of the testing datasets.

| Testing set | | RMSE | MAE | MAPE | $R^2$ |
|---|---|---|---|---|---|
| New developed models | Decision tree | 8.84 | 5.65 | 15.66 | 0.83 |
| | Ridge regression | 16.50 | 12.01 | 33.31 | 0.41 |
| | Support vector regressor | 9.13 | 5.58 | 15.10 | 0.82 |
| | Artificial neural network | 9.93 | 6.70 | 18.45 | 0.79 |
| | Random forest | 7.61 | 4.49 | 12.14 | 0.87 |
| | Gradient boosting | 6.67 | 4.27 | 11.43 | 0.90 |
| | LightGBM | 7.02 | 4.32 | 11.50 | 0.89 |
| | XGBoost | 6.68 | 3.97 | 10.60 | 0.90 |

In the testing dataset, gradient boosting and XGBoost exhibited the most superior results. Gradient boosting attained the minimal RMSE of 6.67, although XGBoost gave the lowest MAE of 3.97 and MAPE of 10.60. Both models achieved the maximum R^2 value of 0.90, and their scatter plots indicate that the majority of points are situated close to the ideal prediction line and within the ±10% error margins. Their residual plots further validate consistent predictions with few and equitable mistakes. Fig. 6 further highlights this superior testing performance by showing that boosting-based models maintain comparatively lower error metrics and higher R^2 than the remaining models. LightGBM exhibited commendable performance (R^2 = 0.89), while random forest demonstrated a little decline in testing accuracy relative to training, indicating moderate overfitting. The decision tree, SVR, and ANN yielded satisfactory although inferior accuracy, whereas ridge regression exhibited the poorest testing performance (R^2 = 0.41).

The results indicate that gradient boosting and XGBoost were the most dependable models for forecasting compressive strength, due to their robust generalization capability and consistent predictive performance.

Table 5 Comparison of the present study with the existing research studies.

| Research Study | Dataset Size | Inputs | Outputs | Model(s) | Splitting | RMSE | MAE | R2 | MAPE (%) |
|---|---|---|---|---|---|---|---|---|---|
| **Present Study** | 1204 | 17 Inputs<br>Cement, Fine aggregate, Coarse aggregate, Water, River sand, Silica Fume, Superplasticizer, W/B, Type of fibers, Fibers%, Aspect ratio, Temperature, Curing age, Specimen width, Specimen length, Specimen height | Compressive Strength | Decision tree<br>Ridge regression<br>Support vector regressor<br>Artificial neural network<br>Random forest<br>Gradient boosting<br>Light GBM<br>XGBoost | 80<br>20 | 6.78<br>14.81<br>8.14<br>8.65<br>3.57<br>4.09<br>3.71<br>3.22 | 4.02<br>10.88<br>4.32<br>5.68<br>4.49<br>2.54<br>2.05<br>1.61 | 0.91<br>0.58<br>0.87<br>0.79<br>0.98<br>0.97<br>0.97<br>0.90 | 10.46<br>29.81<br>10.84<br>15.13<br>4.78<br>6.85<br>5.33<br>4.01 |
| [55] | 445 | 7 Inputs<br>Cement, Water to Cement ratio, Coarse natural aggregates, Fine natural aggregates, Coarse brick aggregates, Fine brick aggregates, Temperature | Compressive Strength | Random Forest<br>Gradient Boosting<br>XGBoost<br>CatBoost<br>Stacking Ensemble | 80<br>20 | 5.36<br>4.52<br>4.82<br>4.51<br>4.38 | 4.09<br>3.32<br>3.57<br>3.31<br>3.23 | 0.881<br>0.915<br>0.903<br>0.915<br>0.919 | _<br>_<br>_<br>_<br>_ |
| [56] | 500 | 8 Inputs<br>Cement, Water, Silica Fume, Fly Ash, Fine aggregates, Coarse Aggregates, Temperature, Gound Granulated Blast-furnace | Compressive Strength | Gradient Boosting<br>Random Forest<br>XGBoost | 70<br>30 | 8.70<br>5.70<br>6.10 | 7.10<br>5.70<br>4.80 | 0.85<br>0.88<br>0.92 | 13.80<br>11.90<br>8.60 |
| [57] | 2279 | 7 Inputs<br>Concrete type, Aggregate type, water to cement ratio, Heating type, Heating rate, maximum core Temperature, Cooling type, Compressive Strength | Compressive Strength | Random Forest<br>AdaBoost<br>GBoost<br>LGBoost<br>XGBoost<br>Linear regression Model | 70<br>30 | 11.90<br>17.41<br>13.90<br>11.20<br>11.00<br>18.80 | 8.27<br>13.7<br>10.3<br>8.1<br>7.5<br>14.5 | 0.855<br>0.691<br>0.803<br>0.871<br>0.876<br>0.639 | _<br>_<br>_<br>_<br>_<br>_ |
| [48] | 674 | 10 Inputs<br>Silica Fume/Cement, GGBS/Cement, Fly Ash/Cement, Steel fiber content, Specimen aspect ratio, Temperature, Aggregate type, Pre-stressed, Pre-drying treatment, Fast heating | Compressive Strength | XGBoost1*<br>XGBoost2*<br>Random Forest<br>XGBoost3*<br>ANN<br>AdaBoost<br>KNN<br>MARS<br>EC2<br>SVR<br>LR | 75<br>25<br>80:20 (only for ANN) | 8.90<br>10.5<br>10.5<br>11.0<br>9.50<br>12.6<br>12.2<br>12.2<br>15.6<br>14.5<br>16.7 | 7.10<br>7.70<br>8.00<br>8.40<br>8.10<br>9.00<br>9.30<br>9.80<br>12.0<br>11.5<br>13.6 | 0.886<br>0.856<br>0.851<br>0.838<br>0.822<br>0.782<br>0.774<br>0.768<br>0.731<br>0.684<br>0.555 | _<br>_<br>_<br>_<br>_<br>_<br>_<br>_<br>_<br>_<br>_ |
| [58] | 169 | 4 Inputs<br>Nano-clay, Nano-silica, Temperature, Duration | Compressive Strength | XGBoost<br>Random Forest | 80<br>20 | 3.90<br>5.13 | 2.47<br>3.83 | 0.998<br>0.994 | 6.35<br>14.43 |
| [59] | 207 | 9 Inputs<br>Cement, Water, Sand, Gravel, Fly Ash, Superplasticizer, Silica Fume, Nano Silica, Temperature | Compressive Strength | AdaBoost<br>Random Forest<br>DT | 80<br>20 | 11.62<br>11.66<br>14.75 | _<br>_<br>_ | 0.938<br>0.935<br>0.911 | 12.52<br>13.07<br>16.10 |

Continued.

| | | | | | | | | | |
|---|---|---|---|---|---|---|---|---|---|
| [34] | 156 | 4 Inputs<br>Temperature degree, Exposure time, Nano alumina, Carbon nanotubes | Compressive Strength | WCA<br>GA<br>ANN<br>FLM<br>MLR | 70<br>30 | 2.954<br>4.551<br>3.194<br>4.822<br>9.818 | 2.293<br>2.940<br>2.404<br>3.362<br>8.568 | 0.985<br>0.974<br>0.983<br>0.962<br>0.815 | _<br>_<br>_<br>_<br>_ |
| [45] | 207 | 9 Inputs<br>Cement, Water, Fine aggregate, Coarse aggregate, Fly Ash, Superplasticizer, Silica Fume, Nano Silica, Temperature | Compressive Strength | Decision tree<br>Bagging<br>Gradient Bosting<br>ANN | 10-fold cross-validation | 10.79<br>7.81<br>9.24<br>11.03 | 7.54<br>5.65<br>6.93<br>9.15 | 0.83<br>0.90<br>0.88<br>0.82 | _<br>_<br>_<br>_ |
| [60] | 307 | 6 Inputs<br>Length, Diameter, Volume fraction, Temperature, Heating Rate, Cementitious Material | Compressive Strength | SVM<br>GBR<br>KNN<br>ANN<br>Random Forest<br>BR | 70<br>30 | 28.75<br>12.93<br>17.93<br>20.05<br>6.27<br>11.82 | 23.42<br>8.31<br>12.57<br>15.59<br>7.23<br>7.97 | 0.75<br>0.95<br>0.90<br>0.87<br>0.96<br>0.95 | 139<br>37<br>41<br>87<br>28<br>24 |
| [61] | 279 | 8 Inputs<br>Cement, Sand, Silt, Fiber, Fiber length, Fiber tensile strength, Specimens age, Type of Fibers | Compressive Strength | ET<br>XGBoost<br>GBR<br>Random Forest<br>DT<br>LIGHTGBM<br>ADA<br>KNN<br>BR<br>RIDGE<br>LR<br>LAR<br>HUBER<br>OMP<br>EN<br>LASSO<br>LLAR | K-fold cross-validation | 0.490<br>0.798<br>0.869<br>0.901<br>0.953<br>1.049<br>1.148<br>1.231<br>1.508<br>1.511<br>1.513<br>1.513<br>1.523<br>1.579<br>1.626<br>1.634<br>1.997 | 0.189<br>0.507<br>0.580<br>0.586<br>0.591<br>0.700<br>0.869<br>0.819<br>1.133<br>1.135<br>1.137<br>1.137<br>1.151<br>1.206<br>1.239<br>1.257<br>1.596 | 0.944<br>0.891<br>0.790<br>0.778<br>0.743<br>0.699<br>0.632<br>0.579<br>0.387<br>0.383<br>0.382<br>0.382<br>0.375<br>0.318<br>0.286<br>0.279<br>-0.08 | 18.99<br>21.03<br>27.65<br>27.45<br>21.88<br>35.83<br>61.98<br>36.98<br>71.27<br>71.26<br>71.28<br>71.28<br>71.41<br>78.66<br>76.27<br>78.31<br>111.8 |
| [62] | 350 | 21 Inputs<br>Water, Cement, Sand, Coarse aggregate, Silica Fume, Fly ash, Quartz powder, Slag, Steel fiber, Steal fiber diameter, Fiber type, other fiber, other fiber diameter, Chamfer ratio, other fiber aspect ratio, Aspect ratio, Curing age, Heating rate, Target temperature, Initial CS, Steel fiber aspect ratio, Edge length or diameter | Compressive Strength | ANN<br>GPR<br>Decision tree<br>Random Forest<br>XGB<br>CATB | 70<br>30 | 0.141<br>0.145<br>0.140<br>0.114<br>0.102<br>0.097 | 0.107<br>0.106<br>0.109<br>0.083<br>0.077<br>0.075 | 0.851<br>0.841<br>0.852<br>0.903<br>0.922<br>0.930 | 0.968<br>1.045<br>0.923<br>0.666<br>0.755<br>0.680 |
| [63] | 324 | 4 Inputs<br>Waste granite powder dose, Waster marble powder, temperature, Duration | Compressive Strength | M5P<br>Random Forest<br>XGBoost | 80<br>20 | 1.408<br>0.840<br>0.420 | 1.133<br>0.677<br>0.327 | 0.933<br>0.976<br>0.993 | 4.01<br>2.36<br>1.16 |

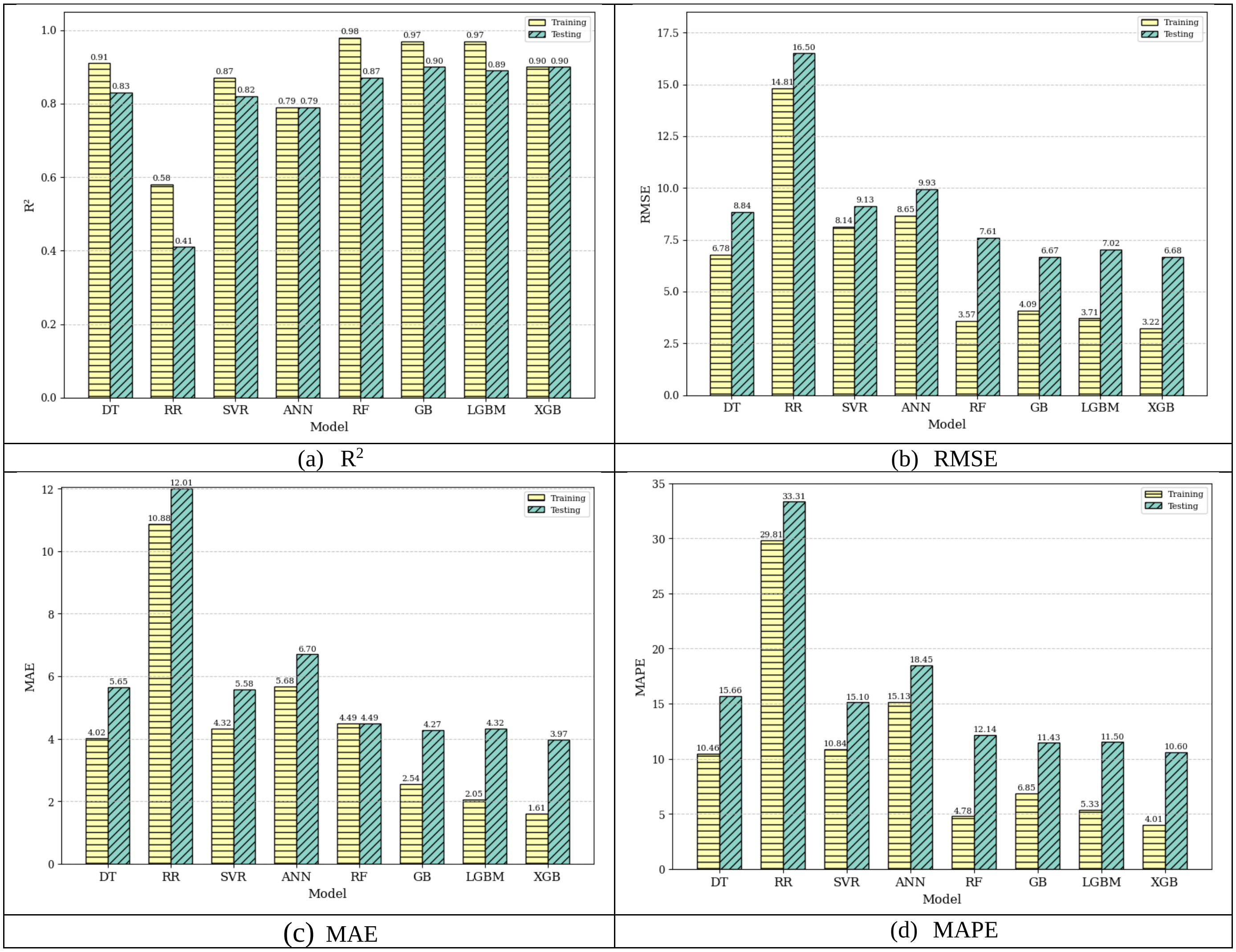

Fig. 6 Performance of the developed models for compressive strength.

### 3.3 SHAP interpretation analysis

SHAP (Shapley Additive Explanations) analysis was conducted to elucidate the predictive behavior of the top-performing models, specifically the gradient boosting (GB) and XGBoost models, as illustrated in Figs. 7–10. The SHAP summary plots and mean absolute SHAP rankings elucidate the significance of the input variables and their impact on predicting compressive strength.

In the GB model (Figs. 7 and 8), temperature emerged as the predominant feature, succeeded by fine aggregate (FAgg), superplasticizer (SP), cement, water-to-binder (W/B) ratio, and water content. The SHAP summary graphic demonstrates that elevated temperature values typically diminished the anticipated compressive strength, whereas decreasing temperature values had a beneficial effect. Conversely, elevated cement and superplasticizer concentrations generally enhanced compressive strength, although increased water-to-binder ratio and water content predominantly exhibited detrimental effects.

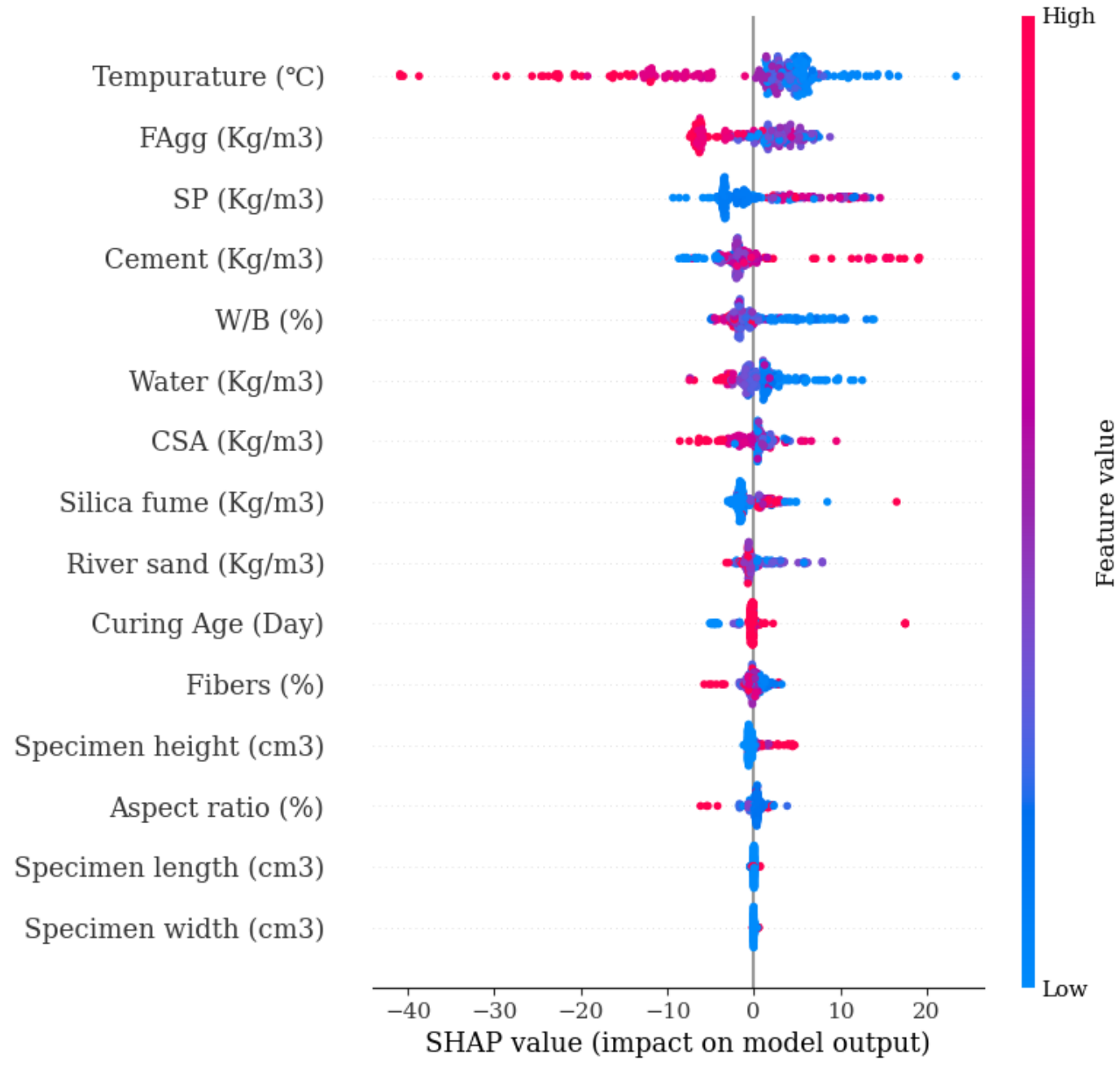

Fig. 7 SHAP interpretation analysis with GB for compressive strength.

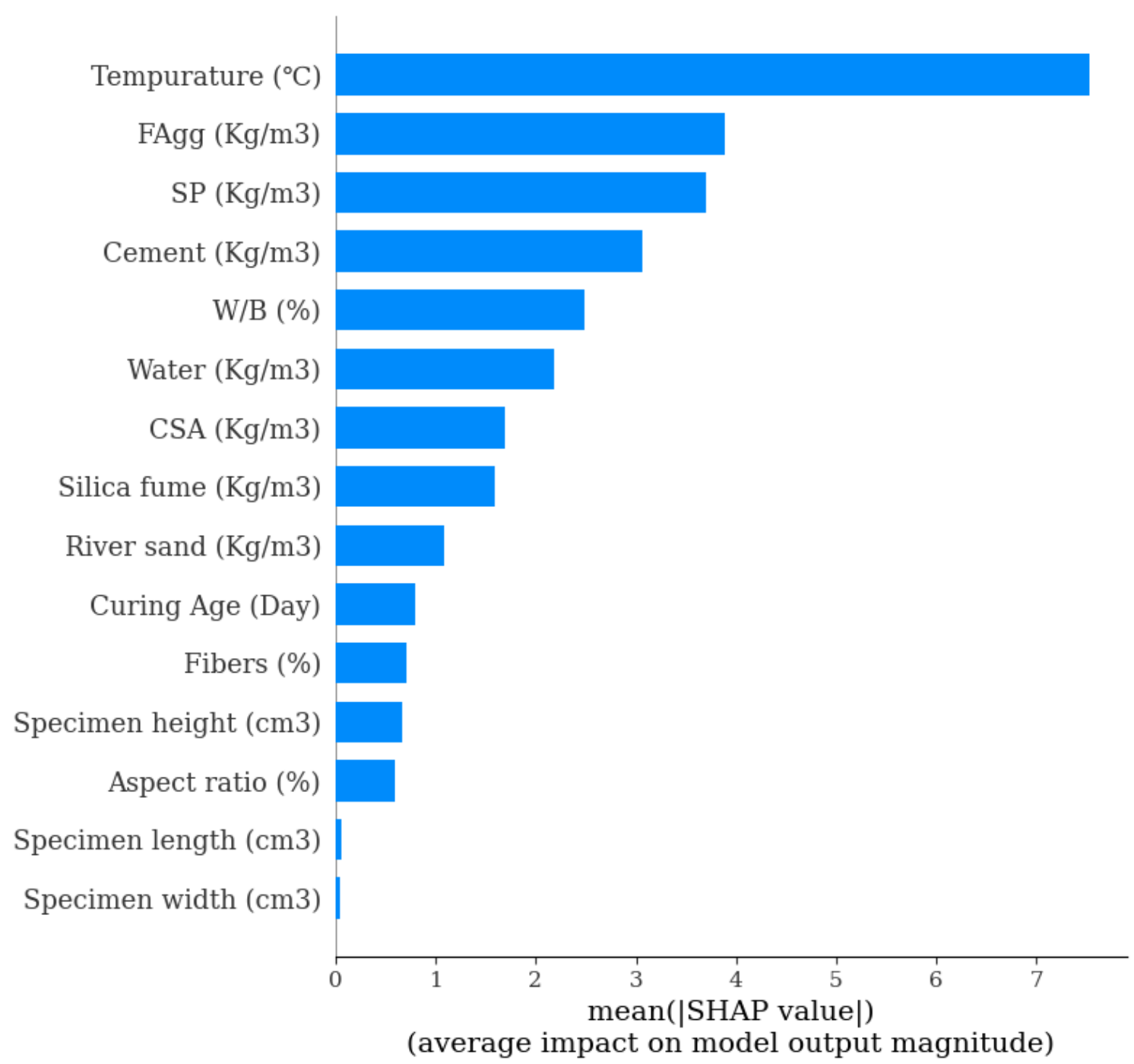

Fig. 8 Feature Importance Ranking (GB) by Mean |SHAP Value|.

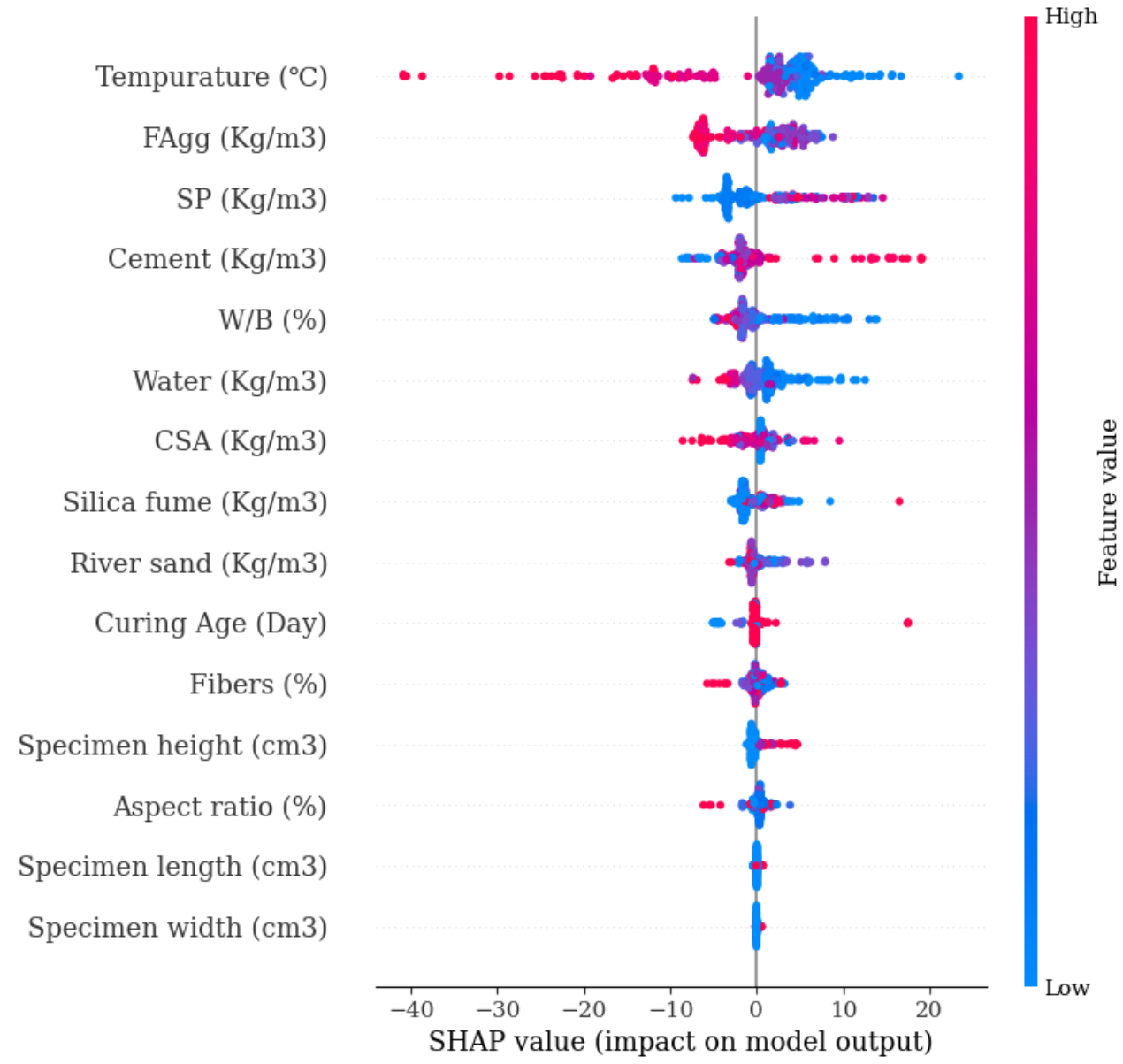

Fig. 9 SHAP interpretation analysis with XGBoost for compressive strength.

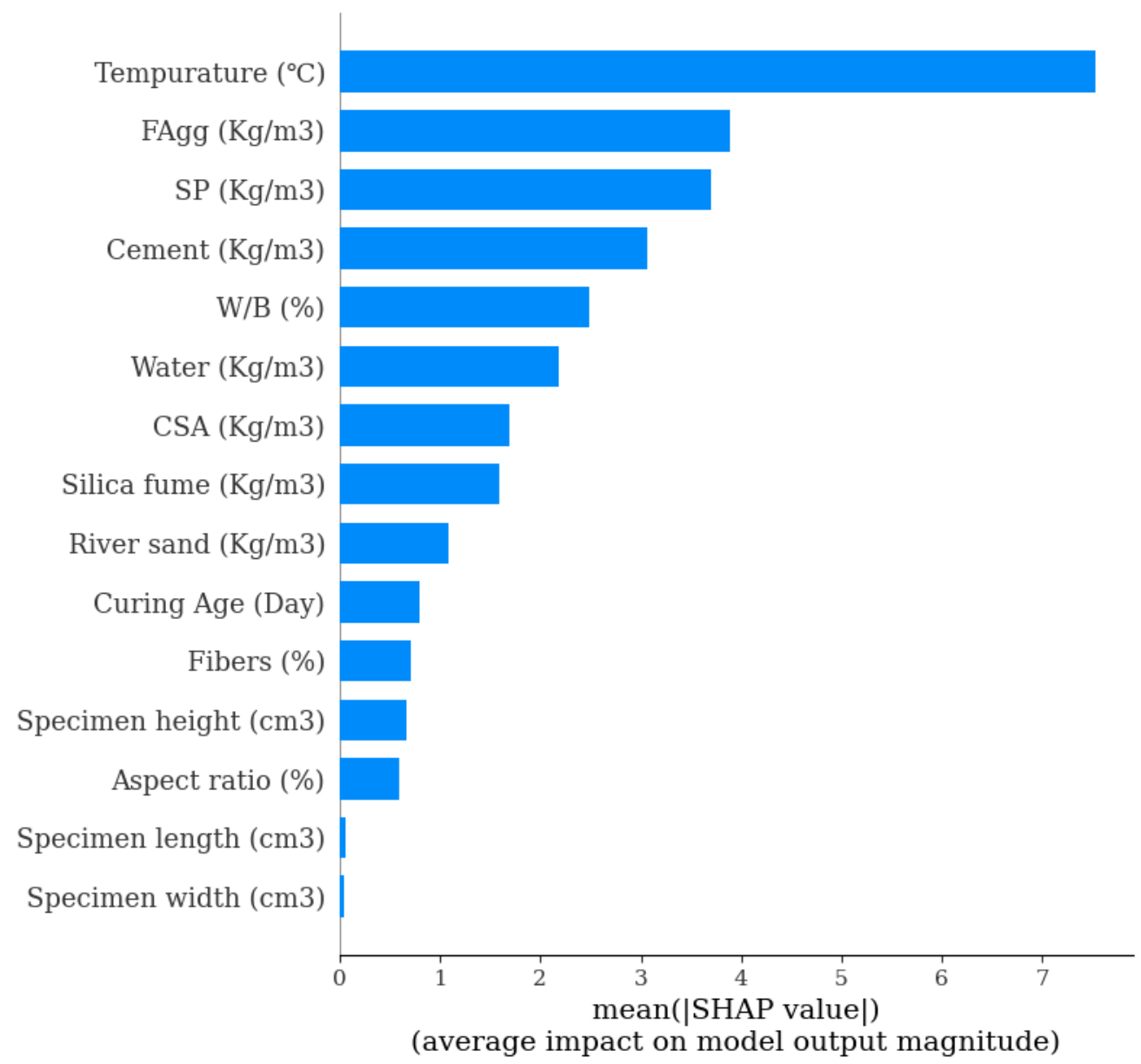

Fig. 10 Feature Importance Ranking (XGBoost) by Mean |SHAP Value|.

A comparable trend was noted with the XGBoost model (Figs. 9 and 10). Temperature was once more identified as the primary influence, succeeded by FAgg, SP, cement, W/B ratio, and water. The strong resemblance between the GB and XGBoost outcomes suggests that both models identified a coherent correlation between the input variables and compressive strength. The SHAP analysis indicates that temperature is the predominant variable, although the mixture proportions also significantly influence compressive strength.

### 3.4 1D and 2D partial dependence plot

Partial dependence visualizations were utilized to analyze the influence of temperature on compressive strength and its interaction with other significant variables. Figure 11 illustrates that the one-dimensional partial dependence plot reveals a negative correlation between temperature and projected compressive strength. The contribution is favorable at low temperatures, but it progressively diminishes and becomes significantly adverse as temperature rises. This outcome aligns well with the SHAP study.

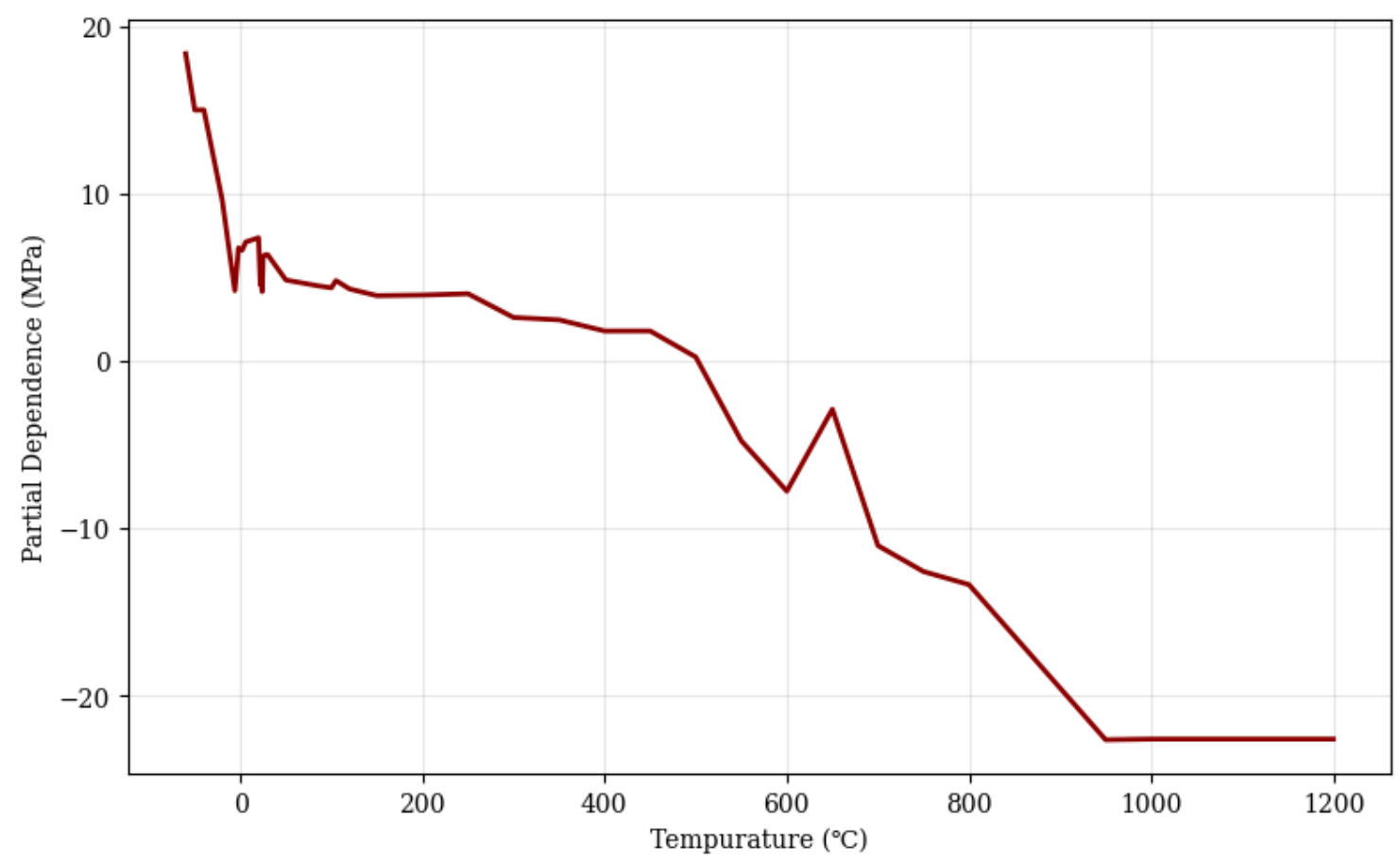

Fig. 11 Effect of temperatures (℃) on predicted compressive strength.

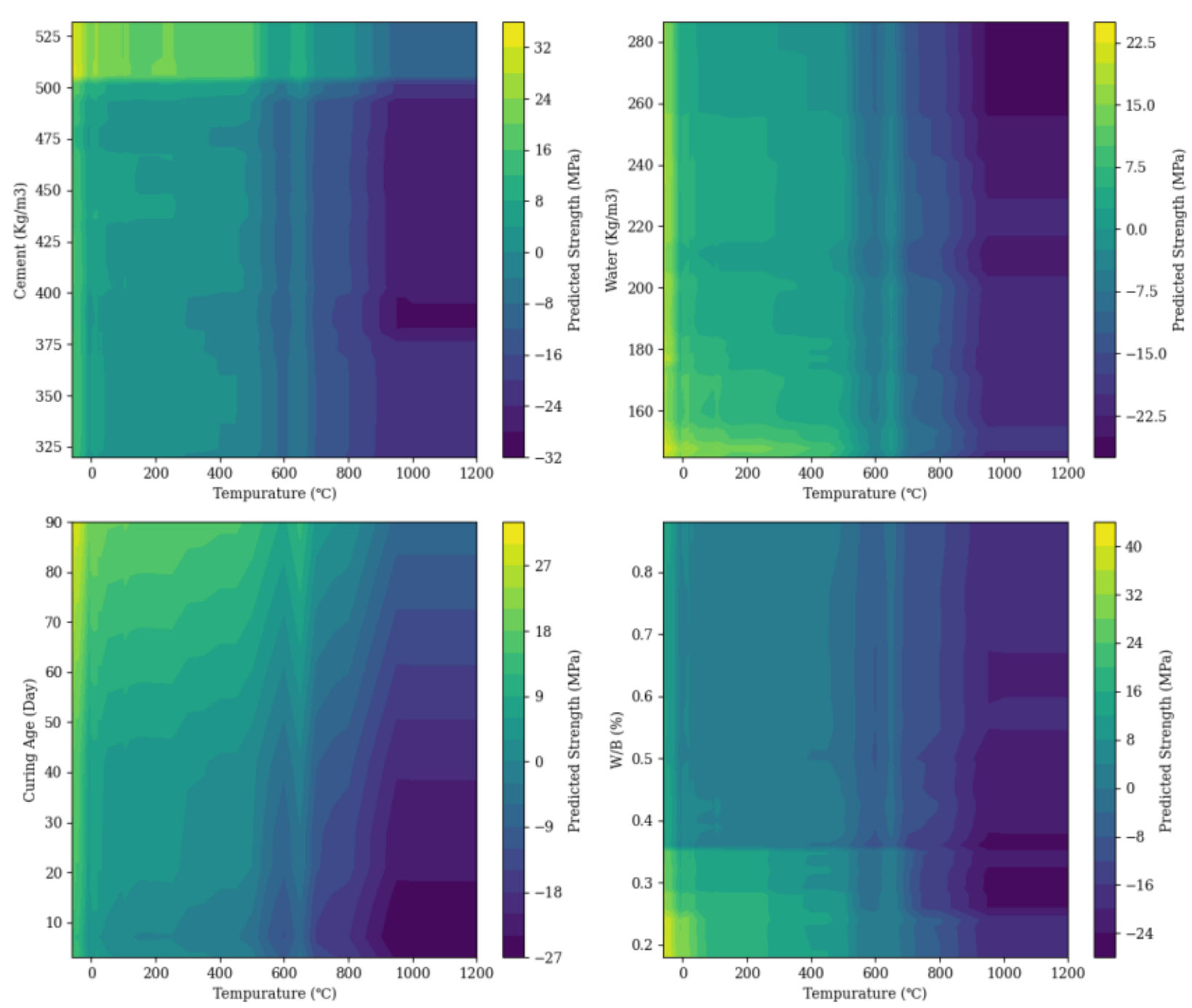

Fig. 12 2D Partial Dependence: Temperature Interaction with Key Mix Parameters.

Figure 12 illustrates the 2D partial dependence plots depicting the interaction between temperature and cement content, water content, curing age, and water-to-binder ratio. In all instances, anticipated compressive strength diminishes with rising temperature, affirming that temperature is the primary determinant. Increased cement concentration and extended curing duration generally enhance compressive strength, particularly at low to moderate temperatures. Conversely, increased water content and a larger water-to-binder ratio diminish the anticipated strength, with this adverse effect becoming more significant at elevated temperatures.

The partial dependence analysis indicates that temperature is the predominant variable, although the impacts of cement, water, curing age, and W/B ratio alter the degree of strength decline.

### 3.5 Comparison of the current study with existing research

The current study's performance was compared with twelve prior research, as detailed in Table 5. The comparison includes dataset size, input variables, utilized ML models, and statistical performance metrics. Model accuracy is contingent not only upon the machine learning technique but also on the dataset size, the quantity of input parameters, and the complexity of the predictive issue. Numerous prior studies utilized comparatively limited datasets and indicated high performance, with certain models attaining R^2 values exceeding 0.90 and exhibiting low RMSE or MAE values. However, same outcomes were frequently achieved with a reduced number of input variables and more constrained datasets. Conversely, certain research utilizing larger datasets demonstrated diminished predictive performance, suggesting that an increased sample size does not invariably yield superior outcomes due to the complexity of the underlying relationships.

This study utilized a substantial dataset of 1204 samples with 17 input variables, evaluating a total of eight machine learning models, which included four individual models and four ensemble models. Among these, gradient boosting and XGBoost exhibited the most superior overall performance. In the testing set, both models attained a maximum R^2 of 0.90, with gradient boosting yielding the lowest RMSE of 6.67 MPa, while XGBoost generated the lowest MAE of 3.97 MPa and MAPE of 10.60%. These values are comparable to several studies presented in Table 5, particularly given the larger dataset and more extensive input space utilized in this research. The comparison demonstrates that the current study attained an effective equilibrium among dataset size, model resiliency, and predictive accuracy. This validates the efficacy of the established ensemble-based models for predicting compressive strength across diverse temperature circumstances.

### 3.6 GUI-based prediction tool

A private graphical user interface (GUI) was created utilizing a Python-based framework and coupled with the two most effective models, specifically gradient boosting (GB) and XGBoost, to offer a functional prediction tool. Figure 13 illustrates that the interface allows users to enter the pertinent mix design and specimen data, choose the desired model, and promptly receive the anticipated compressive strength. The GUI presents the testing performance metrics of each model, including R^2, RMSE, MAE, and MAPE, facilitating a direct comparison between GB and XGBoost. Consequently, the built GUI enhances the practical utility of the suggested ML framework and offers a user-friendly tool for expedited engineering prediction.

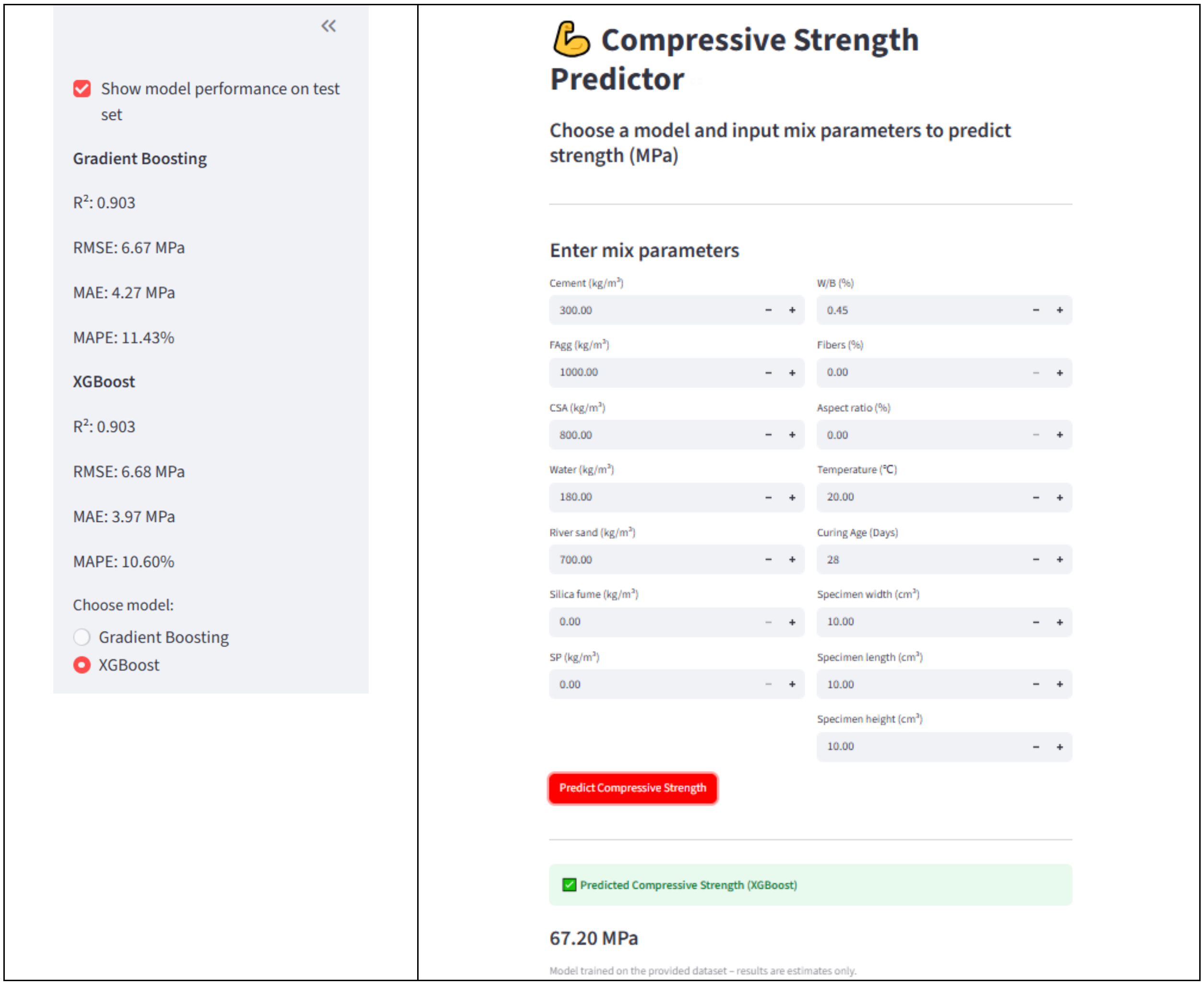

Fig. 13 Python-based GUI developed for compressive strength prediction using GB and XGBoost models.

## 4. Conclusion

This study introduced a machine learning framework for forecasting the compressive strength of fiber-reinforced cementitious composites across different temperature conditions. The study produced an accurate and practicable prediction approach by integrating model creation, performance evaluation, interpretative analysis, comparison with prior research, and GUI application.

1. A total of 1204 datasets comprising 15 input variables were utilised to create and assess eight machine learning models, consisting of four individual models and four ensemble models.
2. The ensemble models demonstrated superior predictive accuracy compared to the individual models, signifying their enhanced capacity to elucidate the intricate nonlinear relationship between the input variables and compressive strength.
3. Among all models, gradient boosting and XGBoost attained the optimal testing performance, each exhibiting the greatest R^2 of 0.90. Gradient boosting yielded the minimal RMSE (6.67 MPa), although XGBoost achieved the lowest MAE (3.97 MPa) and MAPE (10.60%).
4. The SHAP study indicated that temperature was the predominant variable, succeeded by fine aggregate, superplasticizer, cement, water-to-binder ratio, and water content.
5. The one-dimensional and two-dimensional partial dependence graphs further verified that elevated temperatures typically diminished the anticipated compressive strength, whereas the influences of cement, water, curing age, and water-to-binder ratio altered the magnitude of this reduction.
6. The comparison with prior studies indicated that the current study attained competitive predictive accuracy, while utilizing a more extensive dataset and more variety of input factors.
7. A Python-based graphical user interface was created utilizing gradient boosting and XGBoost models, enabling users to input pertinent parameters and promptly receive predictions of compressive strength in a practical and user-friendly manner.
8. The suggested framework offers a precise, interpretable, and practical instrument for predicting compressive strength under temperature-dependent conditions.

## 5. Limitations, future research directions, and practical applications

Despite the robust predictive performance of the generated models, certain limitations persist. The dataset was derived from published studies, and variations in material preparation, curing conditions, and testing methodologies may influence data consistency. The models are anticipated to exhibit greater reliability within the confines of the training data and may demonstrate diminished accuracy for unencountered mixtures or severe temperature scenarios. Future research should concentrate on augmenting the database with more standardized and diverse data, incorporating supplementary material and environmental elements. The experimental validation and prediction of other attributes, including flexural strength, tensile strength, and durability, would enhance the application of the suggested framework. The generated ML models and Python-based GUI serve as efficient instruments for estimating compressive strength, conducting preliminary mix design, and facilitating engineering decision-making in temperature-dependent scenarios.

## CRediT authorship contribution statement

**Jagaran Chakma:** Methodology, Data curation, Data generation, Formal analysis, Validation, Writing–original draft, Investigation. **Zhiguang Zhou:** Supervision, Conceptualization, Investigation, Methodology, Writing – review & editing. **Badhan Chakma:** Data generation, Data curation, Methodology, Visualization, Writing – review & editing. **Priyanta Chakma:** Data generation, Data curation, Writing – review & editing.

## Data availability

The corresponding author will provide the raw data supporting the conclusions of this study upon reasonable request.

## Declaration of competing interest

The authors declare that they have no known competing financial interests or personal relations that may have apparently influenced the work presented in this study.

## Acknowledgements

The authors wish to gratefully acknowledge the support of this work by the National Key Research and Development Program of China under Grant No. 2024YFC3015100.

## Reference

[1] X. Chen, L. Zhang, H. Yu, H. Ma, H. Qiao, L. Yang, H. Fan, and Q. Du, "Mechanical performance and durability evolution of high-performance concrete components exposed to western salt lake environment," Construction and Building Materials, vol. 490, p. 142582, 2025, doi: 10.1016/j.conbuildmat.2025.142582.
[2] J. Lu, "Recent advances in high strength lightweight concrete: From development strategies to practical applications," Construction and Building Materials, vol. 400, p. 132905, 2023, doi: 10.1016/j.conbuildmat.2023.132905.
[3] H. Lou, T. Liu, D. Zou, and Y. Li, "Restoration of fire-damaged high-performance concrete columns by post-fire water-CO2 cyclic curing," Engineering Structures, vol. 353, p. 122253, 2026, doi: 10.1016/j.engstruct.2026.122253.
[4] L. Luo, H. Sun, X. Li, H. Yuan, and X. Cheng, "High-temperature properties and microstructure of a high-performance concrete with recycled aggregates modified by steel fibers and nano-silica," Fire and Materials, vol. 49, no. 7, pp. 1045–1072, 2025, doi: 10.1002/fam.70002.
[5] C. Vu, O. Plé, J. Weiss, and D. Amitrano, "Revisiting the concept of characteristic compressive strength of concrete," Construction and Building Materials, vol. 263, p. 120126, 2020, doi: 10.1016/j.conbuildmat.2020.120126.
[6] S. He, Z. Jiang, H. Chen, Z. Chen, J. Ding, H. Deng, and A. S. Mosallam, "Mechanical properties, durability, and structural applications of rubber concrete: A state-of-the-art review," Sustainability, vol. 15, no. 11, p. 8541, 2023, doi: 10.3390/su15118541.
[7] O. Othman, S. Yehia, N. Qaddoumi, M. Elchalakani, H. Malhat, and W. E. Elemam, "Evaluation of the durability and shielding properties of high-strength concrete incorporating locally available materials and carbon additives," Scientific Reports, vol. 16, no. 1, 2026, doi: 10.1038/s41598-026-37449-3.
[8] S. Bao, S. Wang, S. Wang, X. Tang, and T. Guo, "Effect of fiber type and content on the mechanical properties of high-performance concrete under different saturation levels," Buildings, vol. 15, no. 20, p. 3805, 2025, doi: 10.3390/buildings15203805.
[9] Q. Chen, J. Zhang, L. Zhang, Z. Wang, Z. Wang, T. Zhao, Y. Zhang, Z. Wang, and Z. Wang, "Compressive strength prediction of high-performance concrete: Integrating multi-ingredient influences and mix proportion insights," Construction and Building Materials, vol. 451, p. 138791, 2024, doi: 10.1016/j.conbuildmat.2024.138791.

[10] M. R. Kaloop, D. Kumar, P. Samui, J. W. Hu, and D. Kim, "Compressive strength prediction of high-performance concrete using gradient tree boosting machine," Construction and Building Materials, vol. 264, p. 120198, 2020, doi: 10.1016/j.conbuildmat.2020.120198.
[11] C. Ge, X. Chen, Y. Gong, X. Kong, and F. Chen, "Effect of high temperature on micro-structure and mechanical properties of fiber-reinforced cement-based composites," Crystals, vol. 14, no. 9, p. 778, 2024, doi: 10.3390/cryst14090778.
[12] D. Huang, Y. Wang, Q. Xia, J. Tian, Y. Li, and G. Peng, "Effect of curing temperature on microstructure and durability of concrete: A review," Journal of Engineering, vol. 2026, no. 1, 2026, doi: 10.1155/je/5688760.
[13] Y. Zhang, W. Ren, J. Lei, L. Sun, Y. Mi, and Y. Chen, "Predicting the compressive strength of high-performance concrete via the DR-CatBoost model," Case Studies in Construction Materials, vol. 21, p. e03990, 2024, doi: 10.1016/j.cscm.2024.e03990.
[14] J. Kang, J. Gao, and X. Liu, "Machine learning for predicting the compressive strength of high-performance concrete materials," Multiscale and Multidisciplinary Modeling, Experiments and Design, vol. 8, no. 9, 2025, doi: 10.1007/s41939-025-00981-2.
[15] M. Imani, A. Beikmohammadi, and H. R. Arabnia, "Comprehensive analysis of random forest and XGBoost performance with SMOTE, ADASYN, and GNUS under varying imbalance levels," Technologies, vol. 13, no. 3, p. 88, 2025, doi: 10.3390/technologies13030088.
[16] D. Suenaga, Y. Takase, T. Abe, G. Orita, and S. Ando, "Prediction accuracy of random forest, XGBoost, LightGBM, and artificial neural network for shear resistance of post-installed anchors," Structures, vol. 50, pp. 1252–1263, 2023, doi: 10.1016/j.istruc.2023.02.066.
[17] M. H. R. Sobuz, S. Majumder, M. S. Afrin, A. Alzlfawi, M. K. I. Kabbo, M. Jameel, S. A. Abubakar, and M. Alharthai, "Evaluating the predictive accuracy of supervised machine learning models to explore the mechanical strength of blast furnace slag incorporated concrete," Scientific Reports, vol. 16, no. 1, p. 6712, 2026, doi: 10.1038/s41598-026-36437-x.
[18] D. Sun, P. Zheng, J. Zhang, and L. Cheng, "Optimized gradient boosting framework for data-driven prediction of concrete compressive strength," Buildings, vol. 15, no. 20, p. 3761, 2025, doi: 10.3390/buildings15203761.
[19] A. Najmoddin, H. Etemadfard, A. HosseiniS, and M. Ghalehnovi, "Multi-output machine learning for predicting the mechanical properties of BFRC," Case Studies in Construction Materials, vol. 20, p. e02818, 2023, doi: 10.1016/j.cscm.2023.e02818.
[20] J. Chakma, Z. Zhou, J. Chakma, and C. YuSen, "A data-driven multi-objective approach for predicting mechanical performance, flowability, and porosity in ultra-high-performance concrete (UHPC)," arXiv, 2025, doi: 10.48550/arXiv.2512.21610.
[21] J. Chakma, Z. Zhou, and B. Chakma, "Mechanical strength prediction of steel-polypropylene fiber-based high-performance concrete using hybrid machine learning algorithms," arXiv, 2025, doi: 10.48550/arXiv.2512.21638.
[22] Z. Zhou, J. Chakma, M. A. Hoque, V. Chakma, and A. Ahmed, "Prediction of UHPC mechanical properties using optimized hybrid machine learning model with robust sensitivity and uncertainty analysis," Materials Research Express, vol. 12, no. 8, p. 085703, 2025, doi: 10.1088/2053-1591/adf8c4.
[23] M. Khan, M. F. Javed, H. Alabduljabbar, and F. Ahmad, "Explainable machine learning and ensemble models for predicting fresh properties of self-consolidating concrete," Scientific Reports, vol. 15, no. 1, p. 42288, 2025, doi: 10.1038/s41598-025-21305-x.
[24] J. Chu, B. Guo, T. Zhong, Q. Guan, Y. Wang, and D. Niu, "Explainable machine learning model for predicting compressive strength of CO2-cured concrete," Case Studies in Construction Materials, vol. 22, p. e04589, 2025, doi: 10.1016/j.cscm.2025.e04589.
[25] R. Alyousef, M. F. Rehman, M. Khan, M. Fawad, A. U. Khan, A. M. Hassan, and N. A. Ghamry, "Machine learning-driven predictive models for compressive strength of steel fiber reinforced concrete subjected to high temperatures," Case Studies in Construction Materials, vol. 19, p. e02418, 2023, doi: 10.1016/j.cscm.2023.e02418.
[26] Y. T. Altuncı, "A comprehensive study on the estimation of concrete compressive strength using machine learning models," Buildings, vol. 14, no. 12, p. 3851, 2024, doi: 10.3390/buildings14123851.
[27] R. K. Tipu, S. Bansal, V. Batra, Suman, and G. A. Patel, "Ensemble machine learning models for predicting concrete compressive strength incorporating various sand types," Multiscale and Multidisciplinary Modeling, Experiments and Design, vol. 8, no. 4, 2025, doi: 10.1007/s41939-025-00812-4.
[28] M. U. Gazi, M. T. Hasan, and P. Debnath, "Few-shot meta-learning for concrete strength prediction: A model-agnostic approach with SHAP analysis," AI in Civil Engineering, vol. 4, no. 1, 2025, doi: 10.1007/s43503-025-00064-8.

[29] M. Zhang, Z. Gu, Y. Zhao, Y. Fu, and X. Kong, "Compressive strength prediction of cement base under sulfate attack by machine learning approach," Case Studies in Construction Materials, vol. 21, p. e03652, 2024, doi: 10.1016/j.cscm.2024.e03652.
[30] S. Guo, H. Kou, Y. Bi, and M. Mamlooki, "Predicting the compressive strength of self-compacting concrete by developed African vulture optimization algorithm-Elman neural networks," Scientific Reports, vol. 14, no. 1, p. 20080, 2024, doi: 10.1038/s41598-024-71122-x.
[31] C. Poon, Z. Shui, and L. Lam, "Compressive behavior of fiber reinforced high-performance concrete subjected to elevated temperatures," Cement and Concrete Research, vol. 34, no. 12, pp. 2215–2222, 2004, doi: 10.1016/j.cemconres.2004.02.011.
[32] M. Nithurshan and Y. Elakneswaran, "A systematic review and assessment of concrete strength prediction models," Case Studies in Construction Materials, vol. 18, p. e01830, 2023, doi: 10.1016/j.cscm.2023.e01830.
[33] A. A. Del Savio, D. La Torre Esquivel, E. P. Carbajal, and F. De Andrade Silva, "Experimental and analytical study of temperatures developed by the heat of hydration of high-strength self-compacting mass concrete," Case Studies in Construction Materials, vol. 22, p. e04098, 2024, doi: 10.1016/j.cscm.2024.e04098.
[34] A. M. Zeyad, A. A. Mahmoud, A. A. El-Sayed, A. M. Aboraya, I. N. Fathy, N. Zygouris, P. G. Asteris, and I. S. Agwa, "Compressive strength of nano concrete materials under elevated temperatures using machine learning," Scientific Reports, vol. 14, no. 1, p. 24246, 2024, doi: 10.1038/s41598-024-73713-0.
[35] Y. M. Abbas, G. Fares, and M. I. Khan, "Impact of hot weather conditions on the performance of supplementary cementitious materials concrete," Sustainability, vol. 15, no. 10, p. 8393, 2023, doi: 10.3390/su15108393.
[36] Y. Bu, Z. Cai, J. Peng, Y. Yu, and J. Xu, "Data-driven compressive strength estimation and mix optimization for ultra-high-performance concrete," Journal of Building Engineering, vol. 114, p. 114253, 2025, doi: 10.1016/j.jobe.2025.114253.
[37] A. C. S. Bezerra, P. S. Maciel, E. C. S. Corrêa, P. R. R. Soares, M. T. P. Aguilar Junior, and P. R. Cetlin, "Effect of high temperature on the mechanical properties of steel fiber-reinforced concrete," Fibers, vol. 7, no. 12, p. 100, 2019, doi: 10.3390/fib7120100.
[38] J. Lu, J. Liu, H. Yang, J. Gao, X. Wan, and J. Zhang, "Influence of curing temperatures on the performances of fiber-reinforced concrete," Construction and Building Materials, vol. 339, p. 127640, 2022, doi: 10.1016/j.conbuildmat.2022.127640.
[39] A. H. Akca and N. Ö. Zihnioğlu, "High performance concrete under elevated temperatures," Construction and Building Materials, vol. 44, pp. 317–328, 2013, doi: 10.1016/j.conbuildmat.2013.03.005.
[40] A. Lau and M. Anson, "Effect of high temperatures on high performance steel fibre reinforced concrete," Cement and Concrete Research, vol. 36, no. 9, pp. 1698–1707, 2006, doi: 10.1016/j.cemconres.2006.03.024.
[41] T. M. Pham, H. Hyde, M. K. Kaung, Y. Zhuge, D. T. Tran, D. Vlietstra, and T. M. Tran, "Comparative impact behaviours of ultra high performance concrete columns reinforced with polypropylene vs steel fibres," Defence Technology, vol. 40, pp. 138–153, 2024, doi: 10.1016/j.dt.2024.04.016.
[42] M. Adamu, Y. E. Ibrahim, and H. Alanazi, "Evaluating the influence of elevated temperature on compressive strength of date-palm-fiber-reinforced concrete using response surface methodology," Materials, vol. 15, no. 22, p. 8129, 2022, doi: 10.3390/ma15228129.
[43] M. Tawfik, A. El-Said, A. Deifalla, and A. Awad, "Mechanical properties of hybrid steel-polypropylene fiber reinforced high strength concrete exposed to various temperatures," Fibers, vol. 10, no. 6, p. 53, 2022, doi: 10.3390/fib10060053.
[44] H. Tanyildizi, "Prediction of the strength properties of carbon fiber-reinforced lightweight concrete exposed to the high temperature using artificial neural network and support vector machine," Advances in Civil Engineering, vol. 2018, no. 1, 2018, doi: 10.1155/2018/5140610.
[45] A. Ahmad, K. A. Ostrowski, M. Maślak, F. Farooq, I. Mehmood, and A. Nafees, "Comparative study of supervised machine learning algorithms for predicting the compressive strength of concrete at high temperature," Materials, vol. 14, no. 15, p. 4222, 2021, doi: 10.3390/ma14154222.
[46] J. Liang, K. Liu, C. Wang, X. Wang, and J. Liu, "Mechanical properties of steel fiber-reinforced rubber concrete after elevated temperature," Scientific Reports, vol. 15, no. 1, p. 6965, 2025, doi: 10.1038/s41598-024-80458-3.
[47] L. Li, Z. Wang, J. Wu, X. Du, H. Wang, and W. Liu, "Comparative study on the dynamic mechanical properties of steel fiber reinforced concrete at high temperatures and after high temperature cooling," Construction and Building Materials, vol. 346, p. 128448, 2022, doi: 10.1016/j.conbuildmat.2022.128448.
[48] S. Li and J. R. Liew, "Experimental and data-driven analysis on compressive strength of steel fibre reinforced high strength concrete and mortar at elevated temperature," Construction and Building Materials, vol. 341, p. 127845, 2022, doi: 10.1016/j.conbuildmat.2022.127845.

[49] G. Premkumar and S. S. Selvan, “Prediction of mechanical properties of steel fibre-reinforced concrete under elevated temperature using artificial neural network techniques (ANN),” Frontiers in Built Environment, vol. 11, 2025, doi: 10.3389/fbuil.2025.1610115.
[50] M. Adamu, K. U. Rehman, Y. E. Ibrahim, and W. Shatanawi, “Predicting the strengths of date fiber reinforced concrete subjected to elevated temperature using artificial neural network, and Weibull distribution,” Scientific Reports, vol. 13, no. 1, p. 18649, 2023, doi: 10.1038/s41598-023-45462-z.
[51] Moradi Tavana, M. Habibi, and M. Amiri, “Effect of water to cement ratio on mechanical properties of FRC subjected to elevated temperatures: Experimental and soft computing approaches,” Heliyon, vol. 10, no. 21, p. e39513, 2024, doi: 10.1016/j.heliyon.2024.e39513.
[52] M. A. Moghadam and R. A. Izadifard, “Prediction of the tensile strength of normal and steel fiber reinforced concrete exposed to high temperatures,” International Journal of Concrete Structures and Materials, vol. 15, no. 1, 2021, doi: 10.1186/s40069-021-00485-6.
[53] I. Lee and J. Choi, “Mechanical properties of concrete, reinforcing steel, and CFRP under low-temperature conditions,” International Journal of Concrete Structures and Materials, vol. 19, no. 1, 2025, doi: 10.1186/s40069-025-00843-8.
[55] M. H. H. Rad and V. Toufigh, “The mechanical properties of recycled brick aggregate concrete at elevated temperatures: Ensemble machine learning models and experimental validation,” Materials Today Communications, vol. 49, p. 114401, 2025, doi: 10.1016/j.mtcomm.2025.114401.
[56] J. Zhou, Q. Tian, A. Ahmad, and J. Huang, “Exploring the resilience of supplementary cementitious materials-based concrete to elevated temperatures via modern computing techniques,” Materials Today Communications, vol. 38, p. 107894, 2023, doi: 10.1016/j.mtcomm.2023.107894.
[57] M. Noman, A. Khattak, Z. Alam, M. Yaqub, and E. N. Farsangi, “Predicting the residual compressive strength of concrete exposed to elevated temperatures using interpretable machine learning,” Practice Periodical on Structural Design and Construction, vol. 29, no. 4, 2024, doi: 10.1061/PPSCFX.SCENG-1536.
[58] H. A. Dahish and M. Alturki, “Predicting the effects of nano additives and elevated temperatures on concrete compressive strength utilizing machine learning,” Buildings, vol. 15, no. 18, p. 3349, 2025, doi: 10.3390/buildings15183349.
[59] M. Ahmad, J. Hu, F. Ahmad, X. Tang, M. Amjad, M. J. Iqbal, M. Asim, and A. Farooq, “Supervised learning methods for modeling concrete compressive strength prediction at high temperature,” Materials, vol. 14, no. 8, p. 1983, 2021, doi: 10.3390/ma14081983.
[60] A. Khan, M. S. Ahmad, and A. A. Siddiqui, “Machine learning methods for prediction of compressive strength of steel fibre reinforced concrete at elevated temperatures: A comparative analysis,” Discover Civil Engineering, vol. 3, no. 1, 2026, doi: 10.1007/s44290-026-00414-0.
[61] H. Bentegri, M. Rabehi, S. Kherfane, T. A. Nahool, A. Rabehi, M. Guermoui, A. A. Alhussan, D. S. Khafaga, M. M. Eid, and E. M. El-Kenawy, “Assessment of compressive strength of eco-concrete reinforced using machine learning tools,” Scientific Reports, vol. 15, no. 1, p. 5017, 2025, doi: 10.1038/s41598-025-89530-y.
[62] X. Yu, C. Chen, J. Xie, S. Dong, K. Dai, Y. Lin, and K. Liu, “Data-driven prediction of compressive strength for ultra-high performance concrete exposed to elevated temperatures,” Materials Today Communications, vol. 42, p. 111518, 2025, doi: 10.1016/j.mtcomm.2025.111518.
[63] I. N. Fathy, H. A. Dahish, M. K. Alkharisi, A. A. Mahmoud, and H. E. E. Fouad, “Predicting the compressive strength of concrete incorporating waste powders exposed to elevated temperatures utilizing machine learning,” Scientific Reports, vol. 15, no. 1, p. 25275, 2025, doi: 10.1038/s41598-025-11239-9.